\documentclass[11pt]{article}
\usepackage{setspace}

\usepackage{diagbox}
\usepackage{authblk}
\usepackage{bm}
\usepackage{graphicx}
\usepackage{amssymb,amsmath}
\usepackage{cite}
\usepackage{changes}
\usepackage{amsmath}
\usepackage{algorithm}
\usepackage{algorithmic}
\usepackage{array}
\usepackage{url}
\usepackage{subfigure}
\usepackage{picinpar}
\usepackage{url}
\usepackage{flushend}
\usepackage[latin1]{inputenc}
\usepackage{colortbl}
\usepackage{color}
\usepackage{soul}
\usepackage{multirow}
\usepackage{pifont}
\usepackage{color}
\usepackage{alltt}
\usepackage[hidelinks]{hyperref}
\usepackage{enumerate}
\usepackage{siunitx}
\usepackage{epstopdf}
\usepackage{pbox}
\usepackage{mathtools}
\usepackage{bm}
\usepackage{amssymb,amsmath}
\usepackage{booktabs}
\usepackage{makecell}
\usepackage{array}
\usepackage{url}
\usepackage{threeparttable}
\usepackage{color}
\usepackage{multirow}
\usepackage{longtable}
\usepackage{hyperref}
\hypersetup{hidelinks,
	colorlinks=true,
	allcolors=black,
	pdfstartview=Fit,
	breaklinks=true}
\oddsidemargin
-0.2in \topmargin -.60in
\begin{document}
\begin{spacing}{2.0}

\title{GAN-MDF: A Method for Multi-fidelity Data Fusion in Digital Twins}

\author[1]{Lixue Liu}
%\author[2]{Xueguan Song}
\author[1,2]{Chao Zhang\thanks{Corresponding author (E-mail: chao.zhang@dlut.edu.cn)}}
\author[3]{Dacheng Tao}
\affil[1]{\small{School of Mathematical Sciences, Dalian University of Technology, Liaoning, 116024, P.R. China.}}
\affil[2]{\small{Key Laboratory for Computational Mathematics and Data Intelligence of Liaoning Province, School of Mathematical Sciences, Dalian University of Technology, Dalian,
116024, P.R. China. %E-mail: chao.zhang@dlut.edu.cn
}}
\affil[3]{\small{JD Explore Academy, JD.com, 100176, P.R. China.
}}
% \footnote{$^*$ Corresponding author.}
\renewcommand\Authands{ and }

%	\thanks{This work is partially supported by the National Key R\&D Program of China (Grant No. 2018YFB1700704 \& 2018YFB1702502) and the National Natural Science Foundation of China (Grant No. 11401076 \& 61473328).}

\maketitle

\begin{abstract}
The Internet of Things (IoT) collects real-time data of physical systems, such as smart factory, intelligent robot and healtcare system, and provide necessary support for digital twins. Depending on the quality and accuracy, these multi-source data are divided into different fidelity levels.
High-fidelity (HF) responses describe the system of interest accurately but are computed costly. In contrast, low-fidelity (LF) responses have a low computational cost but could not meet the required accuracy. Multi-fidelity data fusion (MDF) methods aims to use massive LF samples and small amounts of HF samples to develop an accurate and efficient model for describing the system with a reasonable computation burden. In this paper, we propose a novel generative adversarial network for MDF in digital twins (GAN-MDF). The generator of GAN-MDF is composed of two sub-networks: one extracts the LF features from an input; and the other integrates the input and the extracted LF features to form the input of the subsequent discriminator. The discriminator of GAN-MDF identifies whether the generator output is a real sample generated from HF model. To enhance the stability of GAN-MDF's training, we also introduce the supervised-loss trick to refine the generator weights during each iteration of the adversarial training. Compared with the state-of-the-art methods, the proposed GAN-MDF has the following advantages: 1) it performs well in the case of either nested or unnested sample structure; 2) there is no specific assumption on the data distribution; and 3) it has high robustness even when very few HF samples are provided. The experimental results also support the validity of GAN-MDF.
\end{abstract}


\section{Introduction}

Digital twins are digital replica of physical systems which describe how the systems respond to stimuli from the external environment in a timely manner \cite{wu2021digital}. The actual physical processes provide data through sensors to build digital twins while digital twins contribute to help understand, predict and optimize performances of physical processes, both of which co-evolve together. The development of the Internet of Things (IoT) sensors make it possible for collecting real-time data so as to establish connection between physical models and their according digital counterparts. Jia {\it et al.} \cite{jia2020digital} proposed a digital-twin-enabled intelligent clock skew estimation to achieve a clock synchronization for reducing resource consumption in fast-changing industrial IoT environments. The digital twin was built based on a set of heterogeneous clock outputs with sensors equipped on them. A robot-centered smart DT framework named Terra was established by Mo {\it et al.} \cite{mo2021terra} to facilitate the deployment of robots in challenging environments. A multi-view multi-modality perception module was specially designed to descibe the current states of the environment and the robot accurately. Taking advatange of the continuous monitoring and abnormalities detection, Elayan {\it et al.} \cite{elayan2021digital} implemented a intelligent context-aware healthcare system with digital twin framework for heart disease diagnosis.

In order to establish a digital twin, one main challenge is to realize data fusion. Data from different sensors equipped on the same agents or different all contribute to digital twin modeling \cite{compare2019challenges}. Also data from sensors equipped on highly similar agents in the same scenario can be utilized as references for each other \cite{wu2021digital}. Those multi-source data are classified into various fidelity levels according to their quality and accuracy. The multi-fidelity data fusion (MDF) methods are performed when there are only a small amount of high-fidelity (HF) data that are accurate but generated costly, and massive low-fidelity (LF) data that cannot meet the required accuracy but are easier to be obtained. The methods integrates the two kinds of data to provide an accurate description of the system of interest \cite{fernandez2016review,park2017remarks,peherstorfer2018survey} and has played an important role in solving a variety of engineering problems. Guo {\it et al.} \cite{guo2021robust} attempted to give an accurate and rubust identification of flame frequency response which plays a crucial role in thermoacoustic instability analysis with MDF methods. Zhang {\it et al.} \cite{zhang2021multi} developed an optimization framework on aerodynamic shape design supported by the deep neural network MDF. Brevault {\it et al.} \cite{brevault2020overview} reviewed Gaussian process-based MDF methods and tested them on four aerospace related problems to compare their advantages and disadvantages.

Many state-of-the-art MDF methods belong to the interpolation-based category, {\it e.g.}, co-RBF \cite{durantin2017multifidelity}, H-kriging \cite{han2012hierarchical} and LS-MFS \cite{zhang2018multifidelity}. In these models, the LF response ${\bf y}^L({\bf x})$ is approximated by using the interpolation function fitting LF samples and then the HF response is expressed as the linear combination of the resultant LF approximation and the discrepancy function. The coefficients of the final HF approximation models are obtained by means of some specific optimization processes. Although these methods have a well-defined mathematical interpretation, they usually require some specific assumptions on the LF and the HF data, {\it e.g.}, the linear (or almost linear) relation between HF and LF responses or a nested sample structure\footnote{The word ``nested" means that the inputs of the HF samples belong to the set of inputs of LF samples.} or a specific sample distribution ({\it e.g.} Gaussian). Unfortunately, these assumptions will not always hold in practice. 

Instead of directly imposing assumptions on multi-fidelity data,  some recent works adopted the machine learning methods to capture the relatedness information between HF and LF samples. Shi {\it et al.} \cite{shi2019support} proposed the support vector regression-based method (called co-SVR) for MDF, where the kernel function is utilized to map the discrepancy between HF and LF responses into the high-dimensional (or infinite-dimensional) feature space. Meng and Karniadakis \cite{meng2020composite} extended the classical physics-informed neural network (PINN) to the MDF problems and proposed the PINN for MF problems (called MPINN accordingly). The MPINN model is composed of four sub-networks: the first approximates the LF responses corresponding to the inputs of the HF samples to convert the unnested data structure to the nested one; the second (resp. third) captures the linear (resp. nonlinear) relation between LF and HF responses; and the fourth is designed for some specific PDE problems. Xu {\it et al.} \cite{xu2019risk} designed a hierarchical regression framework HR for MDF problems, where a series of regressors are stacked to extract the
LF features from an input from different views and then the extracted LF features are integrated with the original input to
form a regressor as the approximation of the HF response. 

%Xu {\it et al.} \cite{xu2020hierarchical} designed a hierarchical regression framework for MFS modeling, where a series of regressors are stacked to extract the LF features from an input from different views and then the extracted LF features are integrated with the original input to form a regressor as the approximation of the HF response.  

However, there still remains a gap between these existing methods and the desired one. As addressed above, the interpolation-based models are usually built under some specific assumptions on multi-fidelity data. On the other hand, the training of learning-based models generally needs quite a few of HF samples to reach the desired modeling accuracy. Especially in the high-dimensional scenario, they have a higher requirement for the HF sample size because there could be more weights to be determined \cite{sarle1994neural,drucker1997support}.

\subsection{Generative Adversarial Networks (GANs)}\label{sec:gan}

Generative adversarial networks (GANs), originally proposed in \cite{goodfellow2014generative}, are referred to a class of neural networks that are composed of two sub-networks: the generator and the discriminator, and trained in the manner of minimax game between them. 

\begin{figure*}[htbp]
	\begin{center}
		\includegraphics[scale=0.5 ]{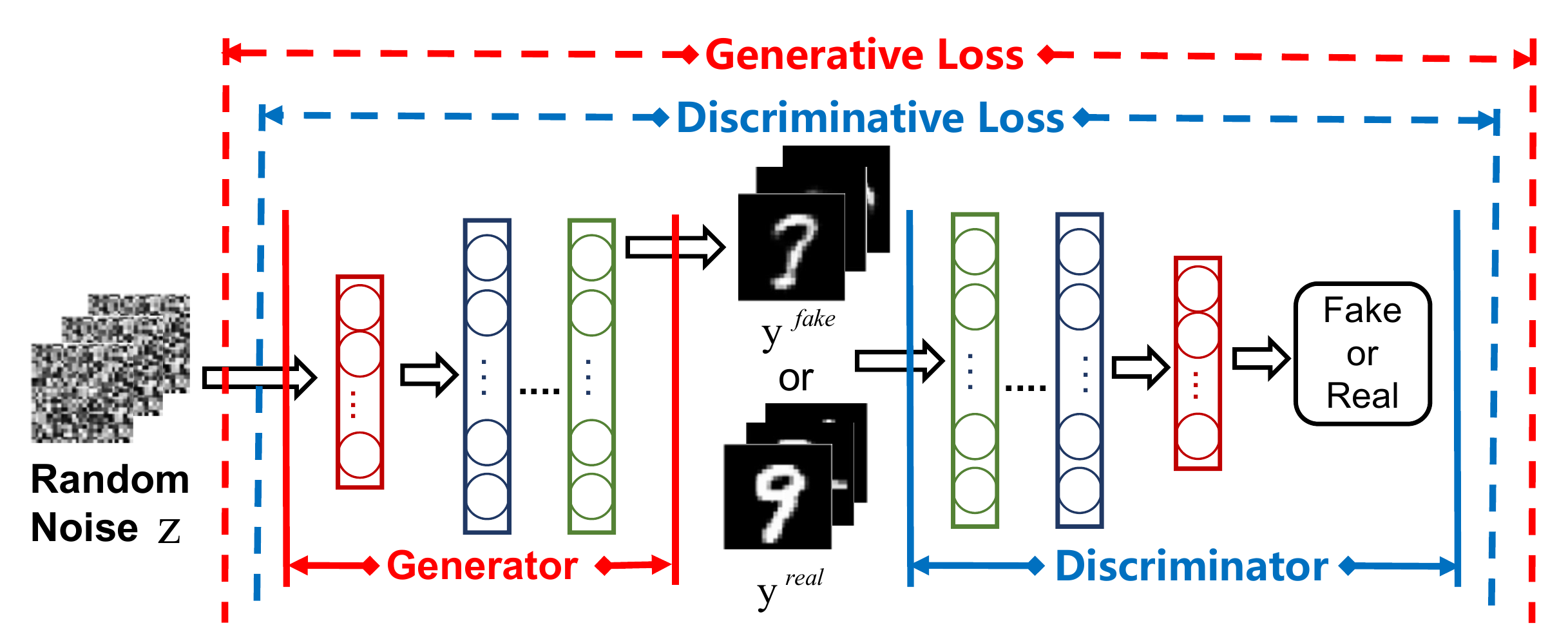}
		%\centering
		\caption{Structure of Original GAN}\label{fig:gan}
	\end{center}
\end{figure*}

The original GAN was designed for the image-generation task, where the inputs can be arbitrary unstructured data ({\it e.g.} random noise) and the outputs are required to be the well-structured ({\it e.g.} image) data \cite{goodfellow2014generative}. As shown in Fig. \ref{fig:gan}, the original GAN is composed of the generator ${\rm G}[\cdot]$ and the discriminator ${\rm D}[\cdot]$. After feeding the unstructured input ${\bf z}$, the generator will produce a fake image ${\rm G}[{\bf z}]$; and the discriminator attempts to identify whether its input is a real image ${\bf y}^{\rm real}$ or the fake image ${\bf y}^{\rm fake}$ produced by the generator. Set ${\rm D}({\bf y})$ as a $[0,1]$-valued function that signifies the probability that the input ${\bf y}$ is a real image ${\bf y}^{\rm real}$. Then, the adversarial training of the original GAN is formalized as the following optimization problem:
\begin{equation}\label{eq:original.GAN}
\min_{\rm G}\max_{\rm D} \bigg\{  \mathbb{E}_{{\bf y}\sim  p_{\rm image}}\big\{\log({\rm D}[{\bf y}])\big\}
+\mathbb{E}_{{\bf z}\sim p_{\bf z}}\big\{\log(1-{\rm D}[{\rm G}[{\bf z}]])\big\}\bigg\},
\end{equation}
where the expectations $\mathbb{E}_{{\bf y}\sim  p_{\rm image}}$ and $\mathbb{E}_{{\bf z}\sim p_{\bf z}}$ are taken on the distributions of the image data and the unstructured data, respectively. In this manner, when the discriminator is trained to be incapable of identifying a generator output or a real image, the distribution of the output provided by the generator will approximate the image-data distribution $p_{\rm image}$ as well as possible, {\it i.e.,} the generator has successfully encoded the complete information of $p_{\rm image}$. More details on GAN and its variants are referred to the survey articles \cite{wang2017generative,pan2019recent}.

Taking advantage of the specific network structure with the adversarial training strategy, many empirical evidences have shown that GANs can approximate the sample distribution more accurately than the traditional machine-learning models obtained by using the same sample set. Thus, GANs have been widely used in many practical application scenarios such as reconstructing high-quality environmental signal according to sensory data from sparsely distributed monitoring sites \cite{kang2020esr}, detecting cyber-attacks in cyber-physical systems with absence of labeled data from novel attacks \cite{de2020intrusion}, improving human gesture recognition performance by generating virtual samples out of a small training sample set \cite{wang2020device}.

\subsection{Overview of Main Results}

In this paper, we propose a generative adversarial network for multi-fidelity data fusion problems (GAN-MDF), where its generator produces the approximation of the HF response of an input; and its discriminator will identify whether the generator output is a real HF response. 

The generator of GAN-MDF is composed of two blocks: the low-fidelity (LF) block and the high-fidelity (HF) block:
\begin{enumerate}[(a)]
	\item The LF block extracts the LF features of an input, and its weights are only trained by using the LF samples and then fixed in the subsequent training process;
	
	\item The HF block, whose weights are trained in the adversarial training process, produces the approximation of the HF response of the input by merging the extracted LF features and the original input together. 
\end{enumerate}
The input of GAN-MDF's discriminator is either a real HF response or a generator output, and the discriminator output is of the $0$-$1$ form. It is expected that the discriminator can identify whether its input is a real HF response or not. Therefore, the labels of the training samples for the discriminator are set in the following way: the real HF responses are labeled as ``true" and the generator outputs are labeled as ``false".

One main challenge in the application of GANs is to guarantee the stability of the adversarial training, which could be influenced by the imbalance between the generator and the discriminator trainings. In the literature, there have been many works on the improvement of GANs' training stability, {\it e.g.,} introducing the specific conditions for controlling GANs' training behavior \cite{mirza2014conditional}; modifying the structures of the original GANs for some particular problems \cite{radford2015unsupervised,choi2018stargan}; 
and redesigning GANs' loss functions \cite{arjovsky2017wasserstein,mao2017least}. In this paper, we introduce the supervised-loss trick to enhance the stability of GAN-MDF's training by refining the generator weights during each iteration. The experimental results validate the trick for the adversarial training of GAN-MDF. Compared with the existing methods, the proposed GAN-MDF has the following advantages:
\begin{enumerate}[(a)]
	\item It is applicable to both nested and unnested samples;
	\item There is no specific requirements on the relatedness between LF and HF responses.
	\item It has a high robustness even when there are very few HF training samples.
\end{enumerate}
%\added{However, when HF and LF functions are (or almost) linearly correlated, the GPR-based methods still perform better.} 
%

%It is noteworthy that the resultant GAN-MFS is originally designed for the bi-fidelity surrogate (BFS) modeling problems, whose samples belong to two fidelity levels. To overcome this limitation, we then introduce a recursive training strategy to extend GAN-MFS to the problems with no less than three fidelity levels. The experimental results also support the effectiveness of the strategy. 

The rest of this paper is organized as follows. In Section \ref{sec:gan-mf}, we give a detailed introduction of GAN-MDF. In Section \ref{sec:exp}, we conduct the numerical experiments to validate GAN-MDF and the last section concludes the paper.

\section{Generative Adversarial Network for Multi-fidelity Data Fusion in Digital Twins (GAN-MDF)}\label{sec:gan-mf}

In this section, we introduce the structure and the adversarial training of GAN-MDF, respectively.

\subsection{Problem Setup}

Let $\{ ({\bf x}_i^L, {\bf y}_i^L) \}_{i=1}^{I_L}\subset \mathbb{R}^{d_1} \times \mathbb{R}^{d_2}$ (resp. $\{ ({\bf x}_j^H, {\bf y}_j^H) \}_{j=1}^{I_H}$) be a set of LF (resp. HF) samples taken from a system with $I_H\ll I_L $, 
{\it i.e.,} the size of LF samples is much larger than that of HF samples. MDF methods aims to use the two sample sets to find a function $f:  \mathbb{R}^{d_1} \rightarrow \mathbb{R}^{d_2}$ such that for any ${\bf x}\in\mathbb{R}^{d_1}$, the function value $f({\bf x})$ can accurately approximate the real HF response of the system at the point ${\bf x}$. 

The main challenge  lies in the lack of enough HF samples to accurately recover the HF model. Inspired by the specific structure and the adversarial training strategy of the original GAN ({\it cf.} Section \ref{sec:gan}), we design a specific generative adversarial network for MDF tasks to exhaustively explore the useful information from the relatively few HF samples. We will show that, benefited from the adversarial strategy, GAN-MDF not only has a lower demand on the size of HF samples but also outperforms the state-of-the-art methods with a higher robustness and stability.

\begin{figure*}[htbp]
	\begin{center}
		\includegraphics[scale=0.5 ]{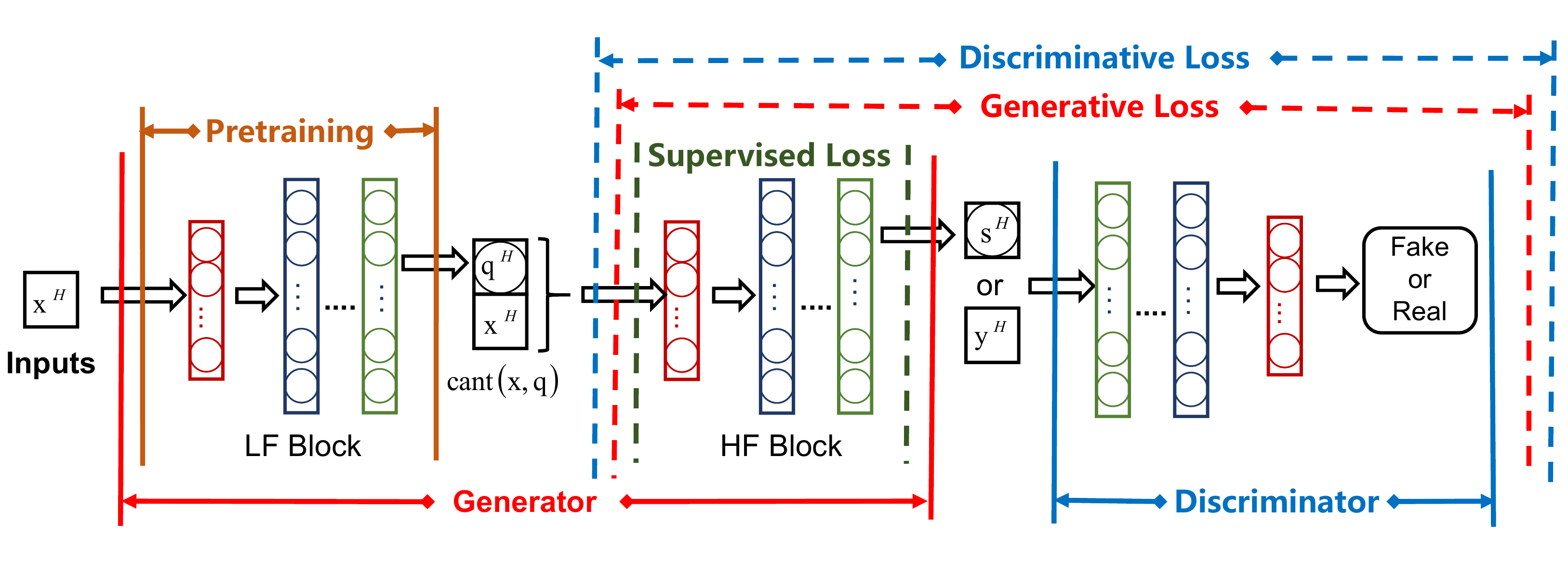}
		%\centering
		\caption{Structure of GAN-MDF}\label{fig:gan-mf}
	\end{center}
\end{figure*}

\subsection{Generator and Discriminator }

As shown in Fig. \ref{fig:gan-mf}, the generator of GAN-MDF is composed of two blocks: the low-fidelity (LF) block and the high-fidelity (HF) block. The LF block is a feed-forward neural network with multiple hidden layers, and its network structure parameters, including node numbers, layer numbers and activation functions, are dependent on the characteristics of specific problems. The following are the candidates for the activation functions in hidden layers of the LF and the HF blocks: 
\begin{enumerate}[(a)]
	\item Sigmoid function
	\begin{equation}\label{eq:sigmoid}
	y=\frac{1}{1+{\rm e}^{-x}};
	\end{equation}
	\item Leaky ReLU function
	\begin{equation}\label{eq:leaky_relu}
	y=\left\{
	\begin{array}{rcl}
	x,  & {x>0}\\
	\alpha x,  & {x \leq 0},
	\end{array} \right.
	\end{equation}
	\item Ricker wavelet function 
	\begin{equation}\label{eq:ricker}
	y=\left(1-2\left(\frac{\pi x}{1000}\right)^2\exp\left(-\left(\frac{\pi x}{1000}\right)^2\right)\right)^2;
	\end{equation}
	
	\item One-dimensional discrete Fourier transform function
	\begin{equation}\label{eq:fourier}
	y(n)=\sum_{k=0}^{M-1}x(k)\cdot {\rm e}^{-\frac{2nk\pi  }{M}i},\quad n=0,1,\cdots,M-1.
	\end{equation}
	
	\item Inverse multi-quadratic function
	\begin{equation}\label{eq: multi-quadratic}
	y = \frac{1}{\sqrt{1+x^2}}.
	\end{equation}
\end{enumerate}
The weights of the LF block are trained by only using the LF samples and then fixed in the subsequent training process. 

Given an input ${\bf x}\in \mathbb{R}^{d_1}$, denote ${\bf q}\in \mathbb{R}^{d_2}$ as the corresponding output of LF block, and the input of HF block is ${\rm cant}({\bf x},{\bf q})\in \mathbb{R}^{d_1+d_2}$, where
\begin{equation*}
{\rm cant}({\bf x},{\bf q}):=(x^1,x^2,\dots,x^{d_1},q^1,q^2,\dots,q^{d_2})^T
\end{equation*}
stands for the concatenation of ${\bf x}$ and ${\bf q}$. The HF block inherits the structure characteristic of LF block, and its weights will be updated in the adversarial training. We denote $\rm{G}[{\bf x}]$ as the output of HF block w.r.t. the sample input ${\bf x}$ and it certainly is the generator output as well.

Moreover, the discriminator structure also inherits the structure characteristic of HF (and LF) block except that the discriminator output is set to be a $0$-$1$ node to identify whether its input is a real HF response (labelled as ``$1$") or not (labelled as ``$0$"). Denote ${\rm D}[G[{\bf x}]]$ as the discriminator output, and its weights will be updated during the adversarial training.

\subsection{Adversarial Training Strategy}

Recalling \eqref{eq:original.GAN}, the generative loss of the original GAN can be of the following form without the logarithmic operation ``$\log(\cdot)$": 
\begin{equation*}%\label{eq:gloss1}
{\cal L}'_{\rm G} := \mathbb{E}_{{\bf x}_1} \big\{{\rm D}[{\bf x}_1]\big\} +   \mathbb{E}_{{\bf x}_2} \big\{1-{\rm D}[{\rm G}[{\bf x}_2]]\big\},
\end{equation*}
where ${\bf x}_1$ and ${\bf x}_2$ stand for the real-data input and the synthesized-data input, respectively. Since the first term does not contain the generator, we simplify the above loss as
\begin{equation*}\label{eq:gloss}
{\cal L}_{\rm G} :=  \mathbb{E}_{{\bf x}} \Big\{1-{\rm D}[{\rm G}[{\bf x}]]\Big\},
\end{equation*}
where ${\bf x}$ is the synthesized-data input. 
By fixing the discriminator weights, the minimization of ${\cal L}_G$ updates the generator weights such that the generator outputs can be identified from the real HF responses by the discriminator. 

Accordingly, the discriminative loss is designed as
\begin{align*}\label{eq:dloss}
{\cal L}_{\rm D} := \mathbb{E}_{{\bf y}^H} &\big\{1-{\rm D}[{\bf y}^H  ]\big\}   +\mathbb{E}_{{\bf x}^H} \big\{ {\rm D}[{\rm G}[{\bf x}^H]]\big\}.
\end{align*}
By fixing the generator weights, the minimization of ${\cal L}_D$ updates the discriminator weights such that it cannot accurately identify whether the generator output is a real HF output.

Different from the traditional GAN training strategy, we introduce the supervised-loss trick to balance the generator and the discriminator trainings in each iteration, and thus stabilize GAN-MDF's training process. Specifically, we introduce the supervised loss w.r.t. the HF samples: 
\begin{equation*}\label{eq:sloss}
{\cal L}_{\rm S} :=   \mathbb{E}_{\left({\bf x}^H,{\bf y}^H\right)}    \Big\{\big\|{\rm G}[{\bf x}^H]-{\bf y}^H\big\|_2^2\Big\},
\end{equation*}
where $\|\cdot\|_2$ stands for the Euclidean norm. During each iteration of the adversarial training, we minimize the loss ${\cal L}_S$ to refine the weights of the generator to prevent divergence from the training of GAN-MDF. The subsequent experimental results demonstrate that the supervised-loss trick plays a key role in the adversarial training of GAN-MDF. Moreover, GAN-MDF will be
trained by using mini-batch Adam method and the workflow of GAN-MDF's training process is sketched 
in Alg. \ref{alg:gan-mf}.

\begin{algorithm}[htbp]
	\renewcommand{\algorithmicrequire}{\textbf{Input:}}
	\renewcommand{\algorithmicensure}{\textbf{Output:}}
	\caption{Adversarial Training of GAN-MF}\label{alg:gan-mf}
	\begin{algorithmic}[1]	
		\REQUIRE LF \& HF samples $\{ ({\bf x}_i^L, {\bf y}_i^L) \}_{i=1}^I$ and $\{ ({\bf x}_j^H, {\bf y}_j^H) \}_{j=1}^{I_H}$, $K$, $\eta_D$, $\eta_G$, $\eta_S$; 
		\ENSURE LF-block weights ${\bf W}_{\rm LF}$, HF-block weights ${\bf W}_{\rm HF}$, discriminator weights ${\bf W}_{\rm D}$;
		\STATE Use $\{ ({\bf x}_i^L, {\bf y}_i^L) \}_{i=1}^I$ to update ${\bf W}_{\rm LF}$ to make the outputs of LF block as accurate as possible;
		\STATE Fix the resultant ${\bf W}_{\rm LF}$ in the following training process; 
		\FORALL{ $k=1,2,\cdots,K$; }
		\STATE Update ${\bf W}_{\rm HF}$ by minimizing the supervised loss ${\cal L}_{\rm S}$: ${\bf W}_{\rm HF}^{(k,1)} = {\bf W}_{\rm HF}^{(k)} - \eta_S\cdot\frac{\partial  {\cal L}_S}{\partial {\bf W}_{\rm HF}};$
		\STATE Update ${\bf W}_{\rm HF}$ and ${\bf W}_{\rm D}$ by minimizing the discriminative loss ${\cal L}_{\rm D}$ respectively: ${\bf W}_{\rm HF}^{(k,2)} = {\bf W}_{\rm HF}^{(k,1)} -  \eta_D\cdot\frac{\partial  {\cal L}_{\rm D}}{\partial {\bf W}_{\rm HF}}$ and ${\bf W}_{\rm D}^{(k,1)} = {\bf W}_{\rm D}^{(k)} -  \eta_D\cdot\frac{\partial  {\cal L}_{\rm D}}{\partial {\bf W}_{\rm D}}$; 
		\STATE Update  ${\bf W}_{\rm HF}$ by minimizing the supervised loss ${\cal L}_{\rm S}$: ${\bf W}_{\rm HF}^{(k,3)} = {\bf W}_{\rm HF}^{(k,2)} -  \eta_S\cdot\frac{\partial  {\cal L}_{\rm S}}{\partial {\bf W}_{\rm HF}};$ 
		\STATE Update ${\bf W}_{\rm HF}$ and ${\bf W}_{\rm D}$ by minimizing the generative loss ${\cal L}_{\rm G}$ respectively: ${\bf W}_{\rm HF}^{(k,4)} = {\bf W}_{\rm HF}^{(k,3)} -  \eta_G\cdot\frac{\partial  {\cal L}_{\rm G}}{\partial {\bf W}_{\rm HF}}$ and ${\bf W}_{\rm D}^{(k+1)} = {\bf W}_{\rm D}^{(k,1)} -  \eta_G\cdot\frac{\partial  {\cal L}_{\rm G}}{\partial {\bf W}_{\rm D}}$; 
		\STATE Update  ${\bf W}_{\rm HF}$ by minimizing the supervised loss ${\cal L}_{\rm S}$: ${\bf W}_{\rm HF}^{(k+1)} = {\bf W}_{\rm HF}^{(k,4)} - \eta_S\cdot \frac{\partial  {\cal L}_{\rm S}}{\partial {\bf W}_{\rm HF}};$
		\ENDFOR
	\end{algorithmic}
\end{algorithm}

\section{Numerical Experiments}\label{sec:exp}

In this section, we conduct the numerical experiments to validate the proposed GAN-MDF in {\it ten} benchmark modeling problems  ({\it cf.} \cite{cutajar2019deep,meng2020composite,tuo2013comment,higdon2002space,cai2017metamodeling,gramacy2009adaptive,an2001quasi,cheng2015trust}). We adopt Latin hypercube sampling method to randomly draw low-fidelity (LF) and high-fidelity (HF) samples in Test-1 to Test-9 according to the functions of these testing problems, respectively. The "Test-10" problem is derived from the real-world dataset of the predicted age-standardised parasite rate for Plasmodium falciparum malaria (PfPR) for global $2$-$10$ year-old children in each year \cite{weiss2019mapping}. The records with the parasite rate larger than $20\%$ in the year of 2000 (resp. 2005) are treated as the LF (resp. HF) data.  Denote $I_L$ and $I_H$ as the LF and the HF training sample size, respectively. This randomly-sampling manner cannot guarantee whether the data structure of these samples are nested or unnested, and thus the experiments are conducted without any assumption on the data structure.

Two candidate data normalization methods are applied including the min-max normalization and standard normalization. A data normalization techinique \cite{enwiki:1012836295} adjust features measured on different scales to a common scale, which is widely used during the data preparation process in machine learning. It not only helps increase the convergence speed and precision of neural networks but avoid gradient explosion as well. The choice of data normalizer depends on the distribution of data. The min-max normalizer linearly rescales every feature by column to the $[0,1]$ interval. Values in each column are transformed in the way of
\begin{equation}\label{eq:minmax}
z = \frac{x-\mbox{min}(x)}{\mbox{max}(x)-\mbox{min}(x)}.
\end{equation}
The standard normalization is performed by mean removal and variance scaling
\begin{equation}\label{eq:standard}
z = \frac{x-\mbox{mean}(x)}{\mbox{stdev}(x)}
\end{equation}
where the mean and standard deviation are computed separately by column. The predictions given by the model will be inversely transformed correspondingly before evaluation.

The loss functions ${\cal L}_S$, ${\cal L}_G$ and ${\cal L}_D$ of GAN-MDF are minimized by using the Adam optimization algorithm with $\min\{32, I_H\}$-size mini-batch. The learning rates are set to be $\eta_L$ for minimizing the the supervised loss during the LF-block training;  $\eta_S$ for minimizing the supervised loss; $\eta_G$ for minimizing the generative loss; and $\eta_D$ for minimizing the discriminator loss, respectively. We note that $\eta_D$ is usually larger than $\eta_G$ so as to enhance the stability of GAN-MDF's training. In Tab. \ref{tab:lr}, we summarize the settings of the learning rates, epoch numbers, activation functions and data normalization methods of GAN-MDF's training in different testing problems.

%\begin{footnotesize}

\begin{table}[htbp]
	\renewcommand\tabcolsep{1pt}
	\centering
	\caption{GAN-MDF's Training Settings for Different BFS Modeling Problems}\label{tab:lr}
	%\resizebox{\linewidth}{!}{
		\begin{threeparttable}
			\begin{tabular}{c|cccccccc}
				\hline
				Problem & $\eta_L$ & $\eta_D$ & $\eta_G$ & $\eta_S$ & $\mbox{epoch}_{LF}$ & $\mbox{epoch}_{HF}$ & \makecell[c]{Activation\\ Functions\tnote{1}} & \makecell[c]{Data\\ Normalizer} \\
				\hline
				Test-1 & 0.03 & 0.002 & 0.001 & 0.05 & 4000 & 350 & \eqref{eq:sigmoid} & None \\
				Test-2 & 0.1 & 0.002 & 0.001 & 0.05 & 4000 & 1500 & \eqref{eq:sigmoid}, \eqref{eq:fourier} & Min-max \\
				Test-3 & 0.1 & 0.002 & 0.001 & 0.05 & 3300 & 1500 & \eqref{eq:sigmoid} & Standard \\
				Test-4 & 0.04 & 0.002 & 0.001 & 0.05 & 5000 & 2000 & \eqref{eq:sigmoid}, \eqref{eq:leaky_relu}, \eqref{eq: multi-quadratic} & Min-max\\
				Test-5 & 0.03 & 0.002 & 0.001 & 0.03 & 4000 & 1100 & \eqref{eq:sigmoid} & Min-max\\
				Test-6 & 0.03 & 0.001 & 0.0005 & 0.003 & 1200 & 1100 & \eqref{eq:sigmoid} & Standard\\
				Test-7 & 0.005 & 0.002 & 0.001 & 0.01 & 1000 & 1000 & \eqref{eq:sigmoid} & None\\
				Test-8 & 0.01 & 0.002 & 0.001 & 0.05 & 500 & 900 &  \eqref{eq:sigmoid} & Min-max\\
				Test-9 & 0.01 & 0.002 & 0.001 & 0.05 & 1500 & 1500 & \eqref{eq:sigmoid}, \eqref{eq:ricker} & Min-max\\
				Test-10 & 0.05 & 0.002 & 0.001 & 0.03 & 4000 & 1000&  \eqref{eq:sigmoid} & Min-max\\
				\hline
			\end{tabular}
			
			\begin{tablenotes}
				\footnotesize
				\item[1] {We mainly concern with the active functions used in the hidden layers of the LF and the HF blocks.}
			\end{tablenotes}
	\end{threeparttable}
%}

\end{table}
%\end{footnotesize}

Given a testing HF sample set $\{({\bf x}^H_n, {\bf y}_n^H) \}_{n=1}^{N_t}$, we adopt the normalized root mean square error (NRMSE) as the criteria of modeling performance:
\begin{equation*}
\mathrm{NRMSE}:=\frac{\sqrt{\sum_{n=1}^{N_t}\|{{\bf y}_n^H}-f^{H}({\bf x}_n)\|^2}}{\sqrt{\sum_{n=1}^{N_t}{\|{\bf y}_n^H\|^2}}}.
\end{equation*}
For each choice of the varying HF sample sizes, the experiment will be repeated {\it ten} times, and the average NRMSEs are recorded as ``Average NRMSEs".

As a comparison, four state-of-the-art MDF methods are considered in the experiments including the co-RBF model \cite{durantin2017multifidelity}, the H-kriging model \cite{han2012hierarchical}, the LS-MFS model \cite{zhang2018multifidelity} and the Hierarchical Regression model (HR) \cite{xu2020hierarchical}. To validate the supervised-loss trick, we then examine the performance of GAN-MF without the supervised loss, and such a network is called as pure-GAN (pGAN) in the experiment. Since the co-RBF model is originally designed for the nested samples, we use the unnested samples to form a nested sample set for training co-RBF model in the following way: 1) we first use LF samples to build a RBF model as the LF model; 2) we then employ the resultant LF model to produce LF responses corresponding to the inputs of HF samples; and 3) we finally treat the pair of each HF input and its corresponding LF response as a new LF sample. The interpolation-based models are not suitable for data normalization in principle, and machine learning-based HR adopts the same data normalizer as GAN-MDF in each problem. Because HR actually provides a modeling framework for MDF, here we employ "ADR-2-LA" as the benchmark method which performed best in the paper \cite{xu2020hierarchical}. The "ADR-2-LA" model use random forrest as basic regressors of Adaboost regression to extract LF features, and the dimension of stacked LF features is reuduced to 2 by PAC learning and the final regressor is chosen as Lasso.   

\begin{figure*}[htbp]
	\begin{minipage}[t]{0.332\linewidth}
		\centerline{\includegraphics[width=\linewidth]{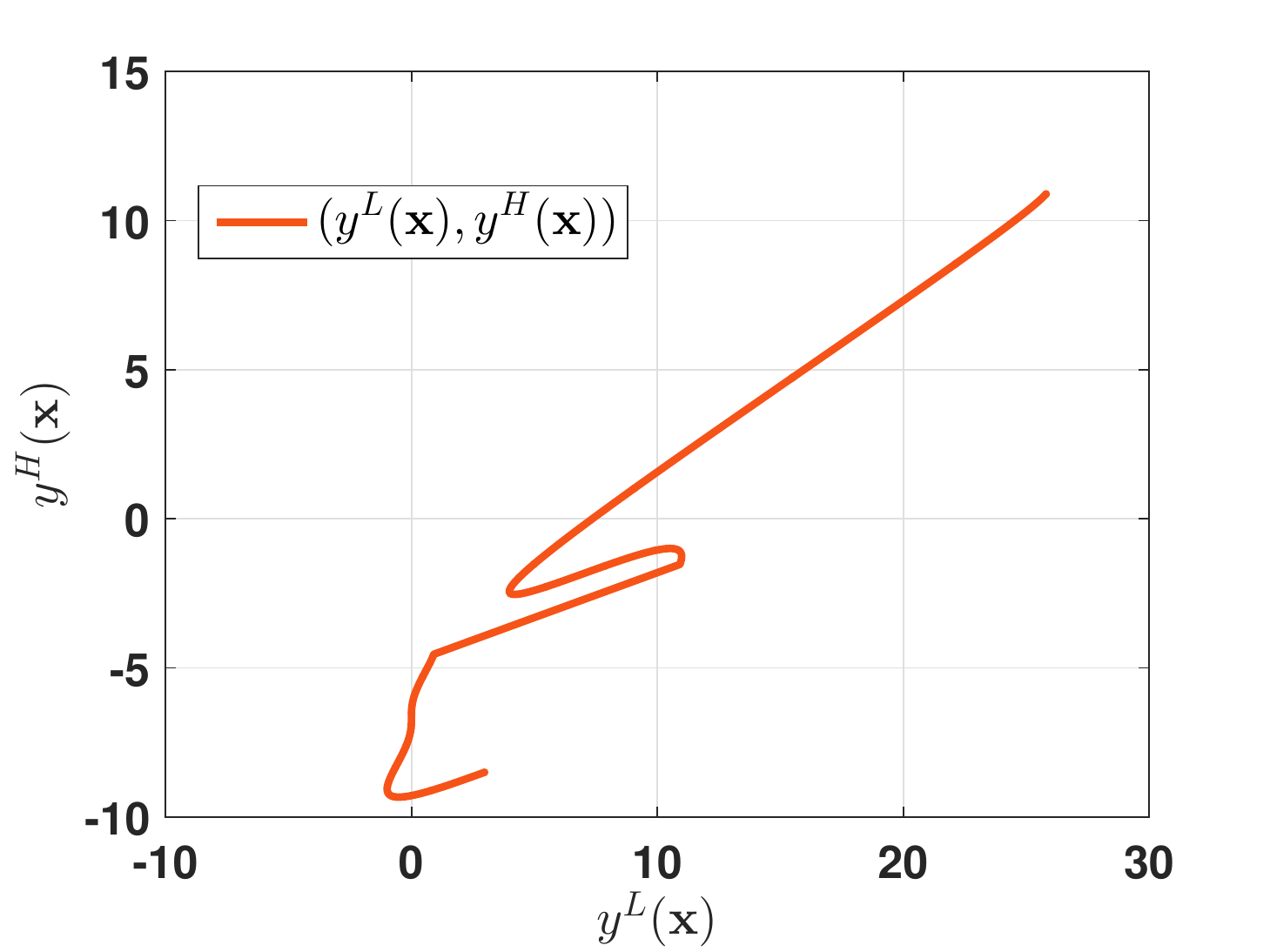}}
		\centerline{(a) Test-1}
	\end{minipage}
	%\hfill	
	\begin{minipage}[t]{0.332\linewidth}
		\centerline{\includegraphics[width=\linewidth]{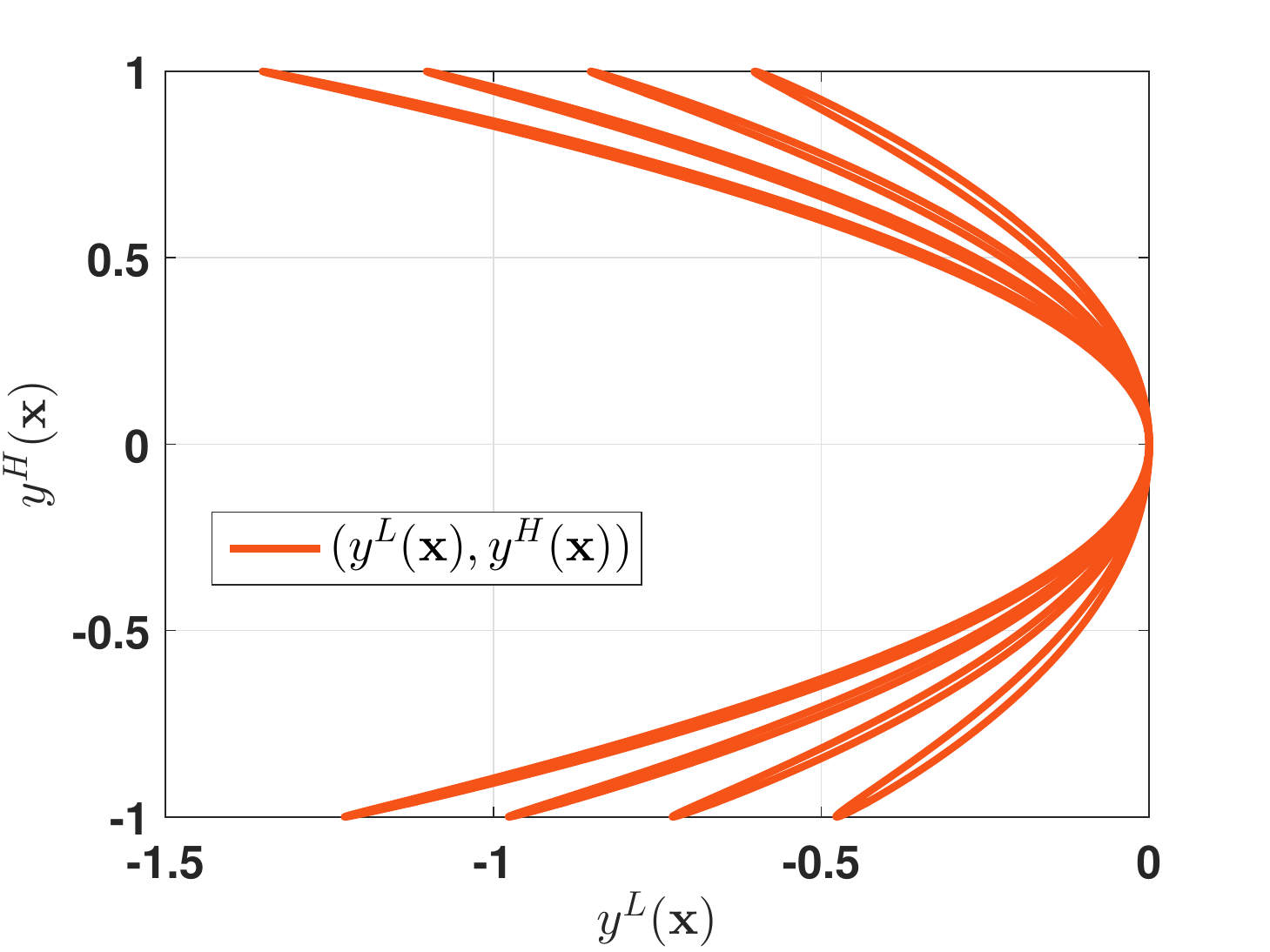}}
		\centerline{(b) Test-2}
	\end{minipage}			
	%\hfill
	\begin{minipage}[t]{0.332\linewidth}
		\centerline{\includegraphics[width=\linewidth]{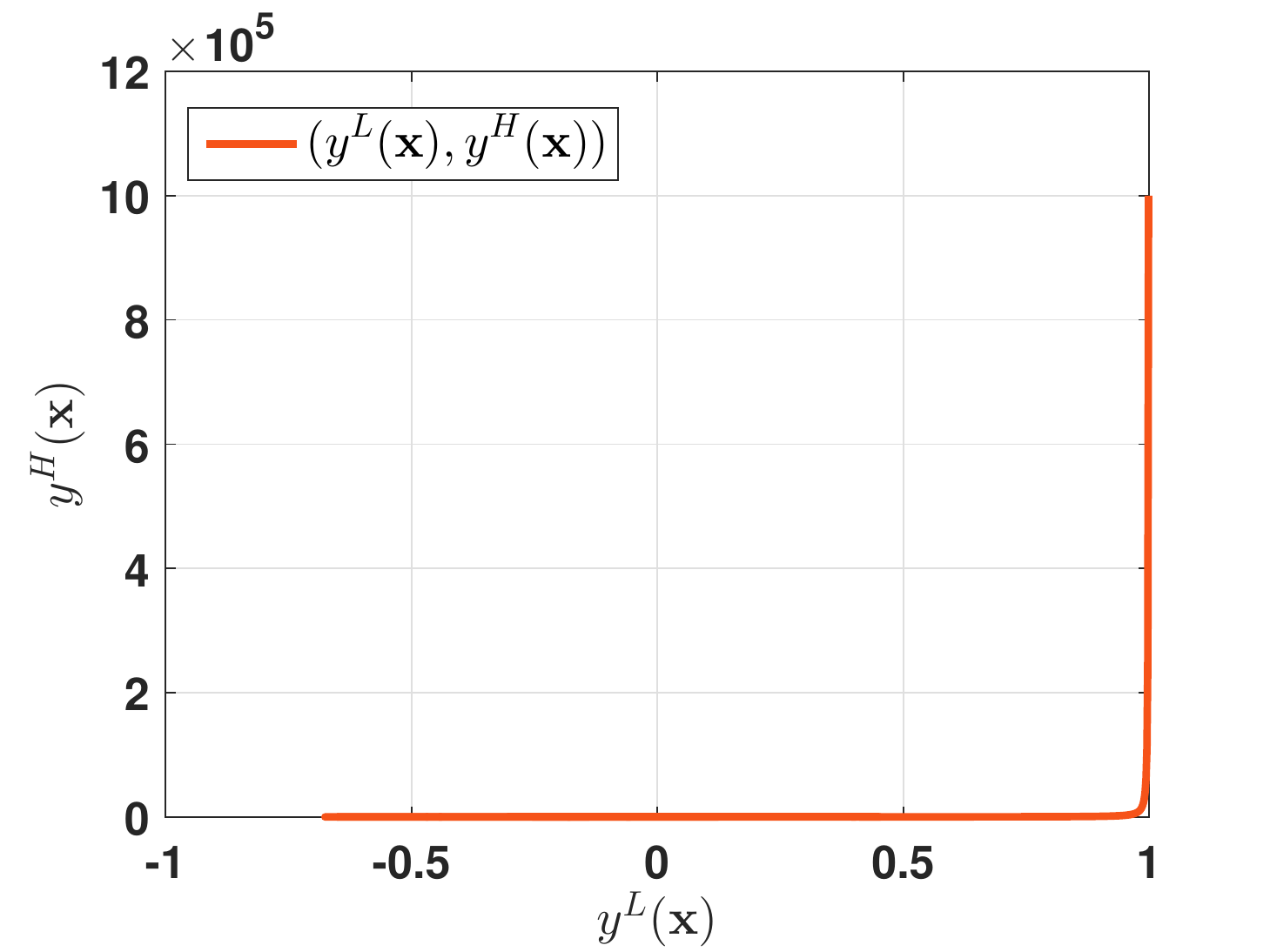}}
		\centerline{(c) Test-3}
	\end{minipage}
	\begin{minipage}[t]{0.332\linewidth}
		\centerline{\includegraphics[width=\linewidth]{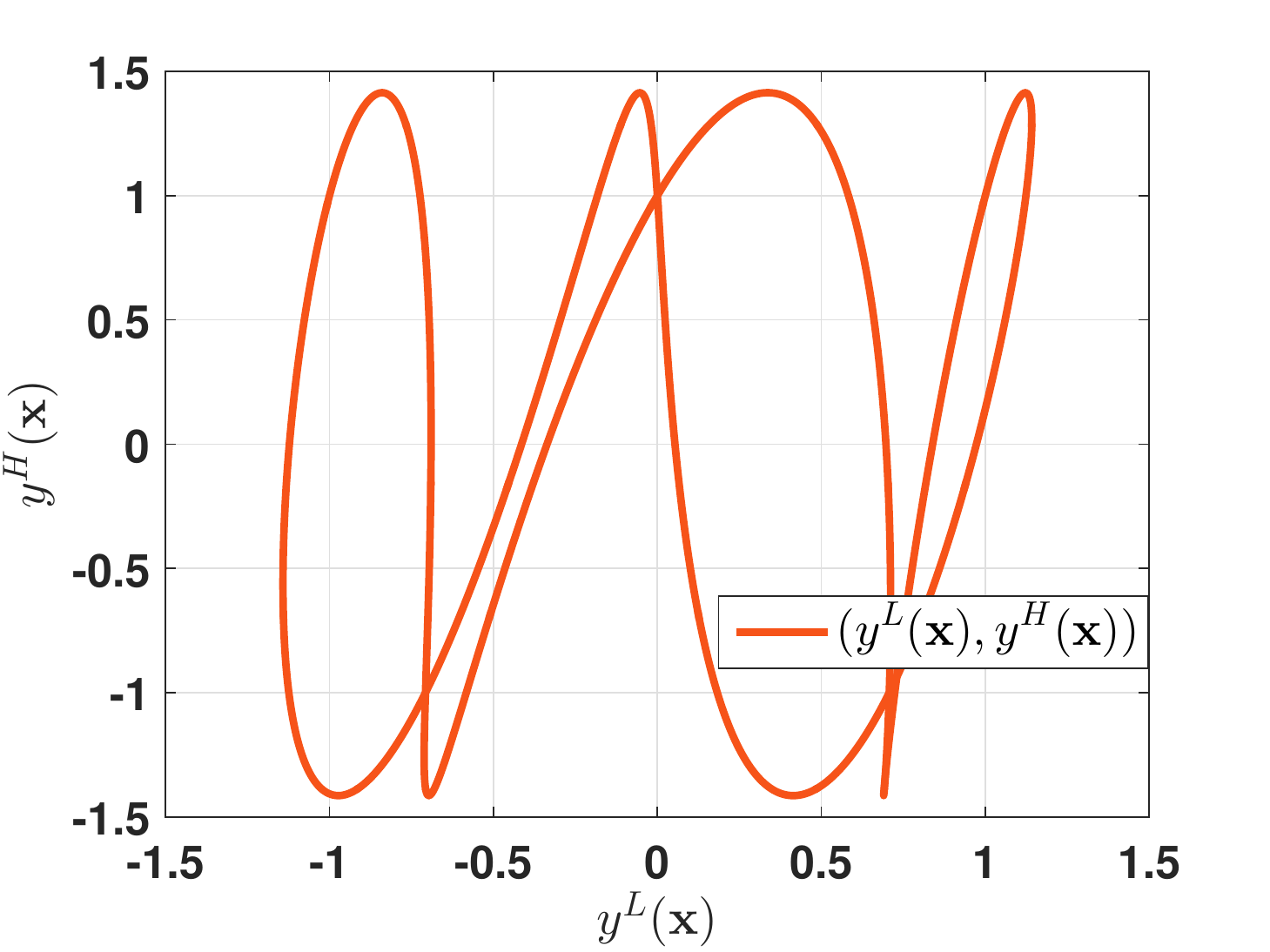}}
		\centerline{(d) Test-4}
	\end{minipage}
	\begin{minipage}[t]{0.332\linewidth}
		\centerline{\includegraphics[width=\linewidth]{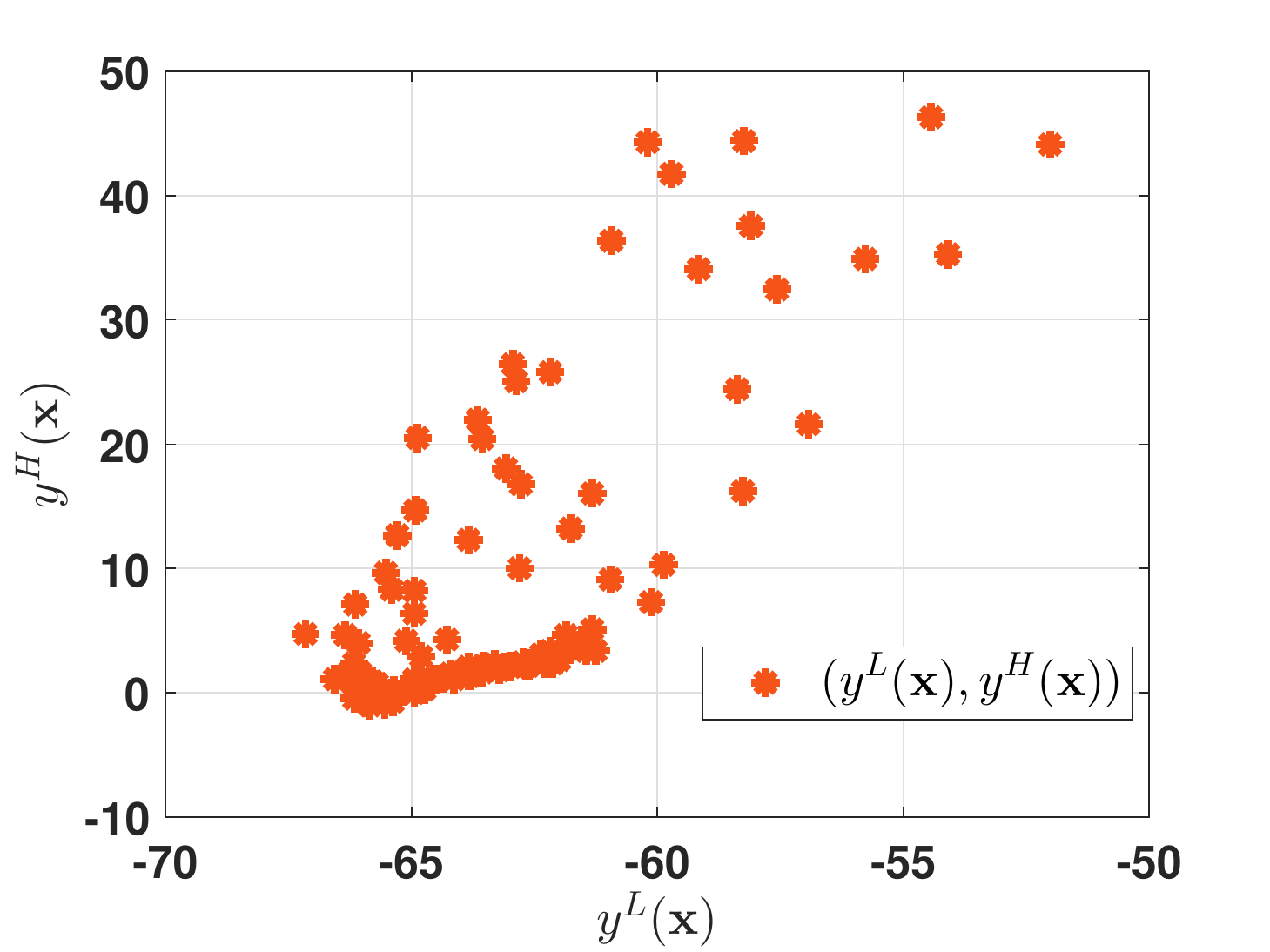}}
		\centerline{(e) Test-5}
	\end{minipage}
		\begin{minipage}[t]{0.332\linewidth}
			\centerline{\includegraphics[width=\linewidth]{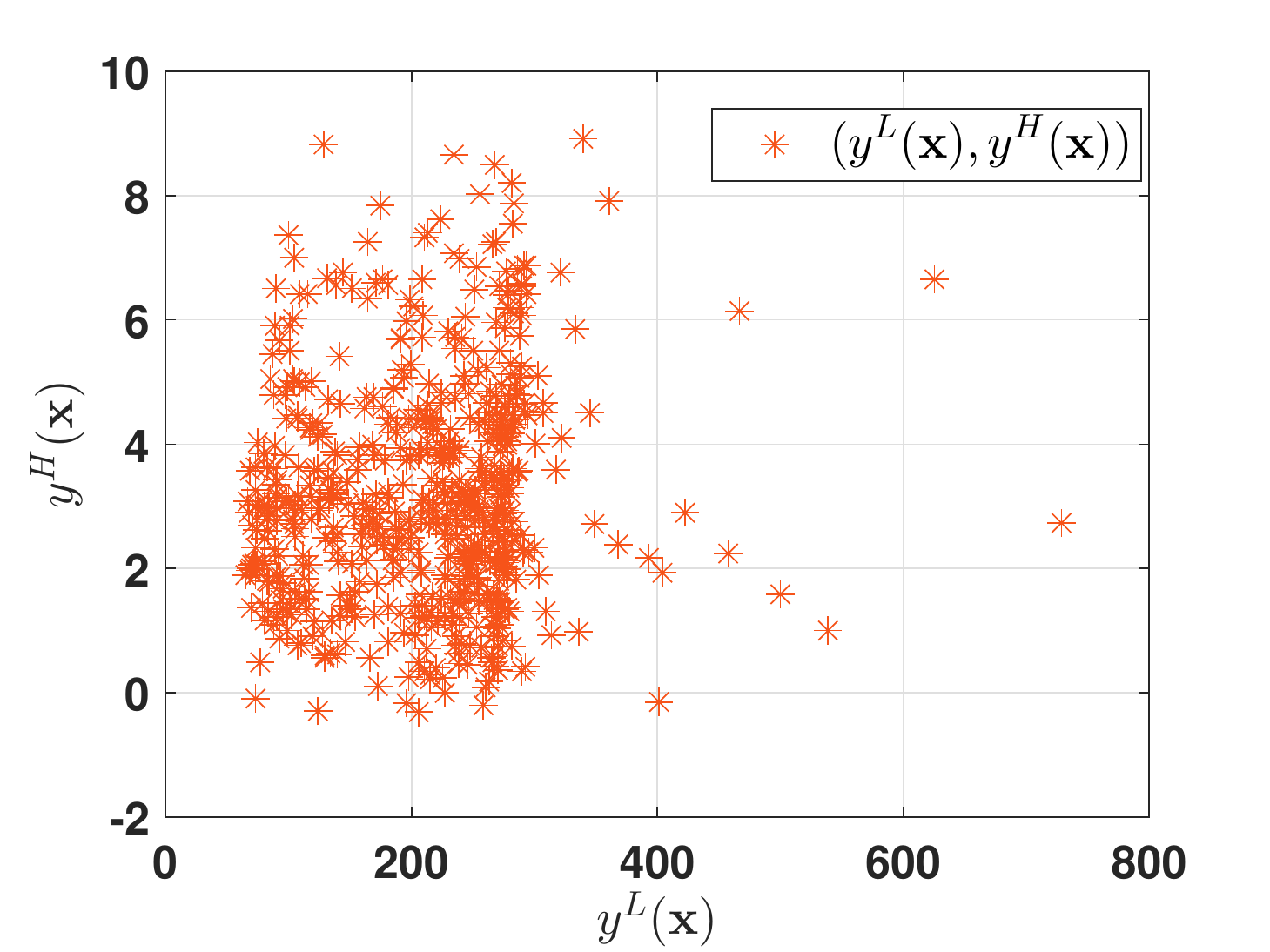}}
			\centerline{(f) Test-6}
		\end{minipage}
		\begin{minipage}[t]{0.332\linewidth}
			\centerline{\includegraphics[width=\linewidth]{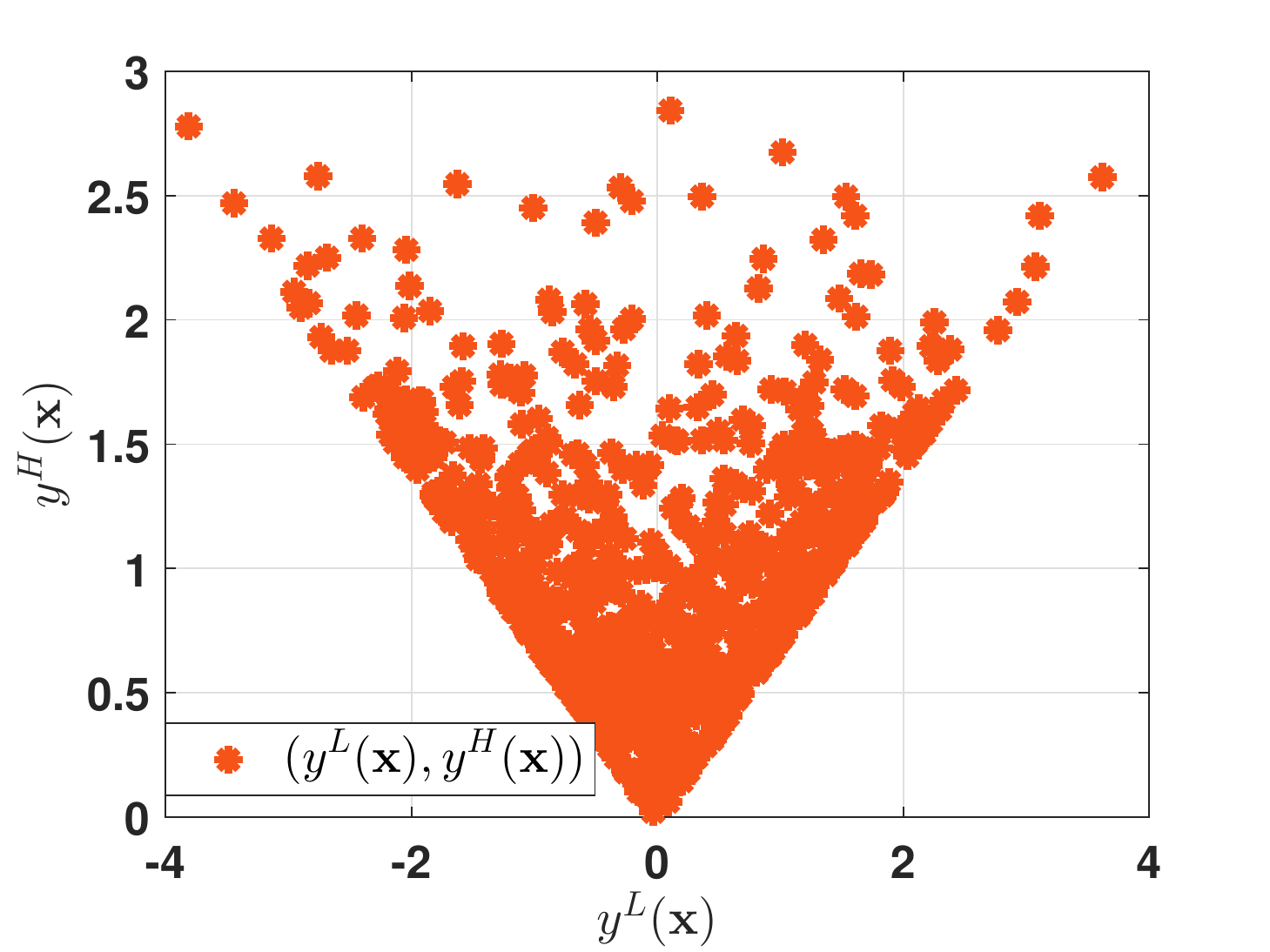}}
			\centerline{(g) Test-7}
		\end{minipage}
		\begin{minipage}[t]{0.332\linewidth}
			\centerline{\includegraphics[width=\linewidth]{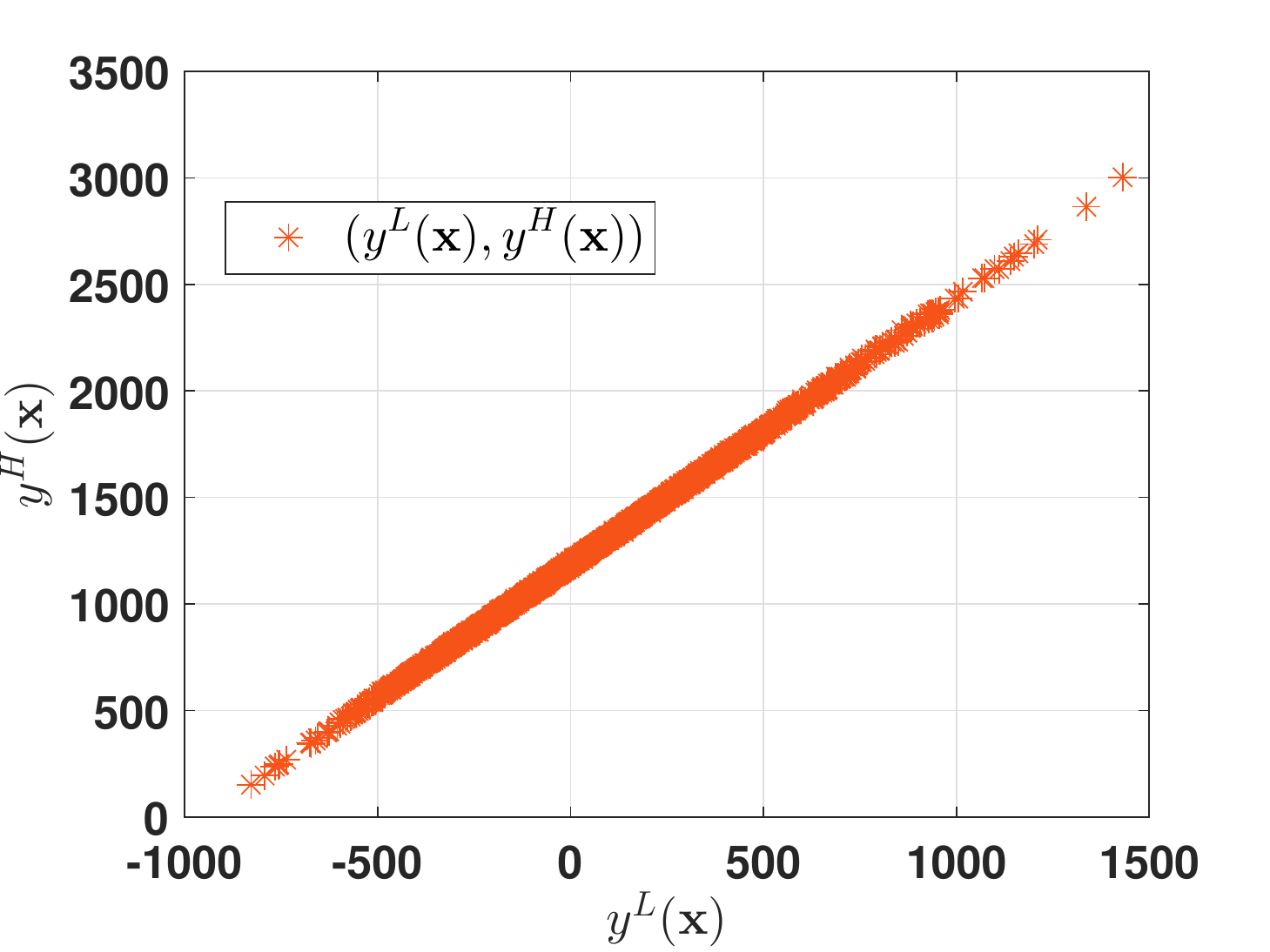}}
			\centerline{(h) Test-8}
		\end{minipage}
		\begin{minipage}[t]{0.332\linewidth}
			\centerline{\includegraphics[width=\linewidth]{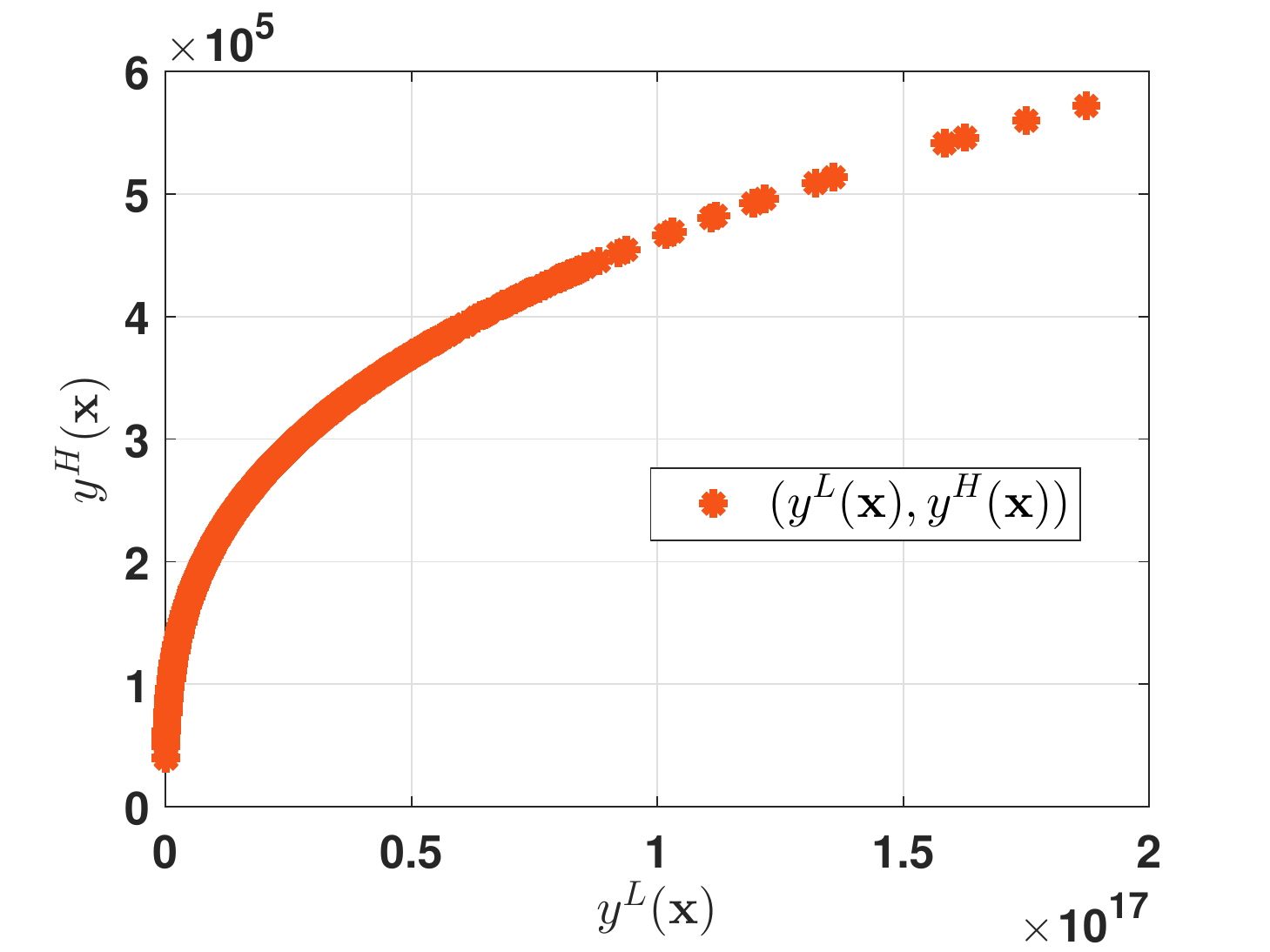}}
			\centerline{(i) Test-9}
		\end{minipage}

	\caption{The functional relations between LF response $y^{L}({\bf x})$ and HF response $y^{H}({\bf x})$ (${\bf x}\in\mathbb{R}^{d_1}$) of benchmark problems}
	\label{fig:bi-corr}
\end{figure*}

In the literature, the correlation coefficient is usually used to measure the correlation between LF and HF responses in the testing problems ({\it cf.} \cite{song2019a}). However, the correlation coefficient actually is the cosine similarity between the centered LF and HF response data, and thus cannot describe the intrinsic relatedness between LF and HF responses. Instead, we draw the points $\{(y^{L}({\bf x}), y^{H}({\bf x}))\}_{{\bf x}\in\mathbb{R}^{d_1}}$ in the plane-coordinate system to characterize the functional relation between LF and HF responses in different testing problems ({\it cf.} Fig. \ref{fig:bi-corr}).

\subsection{Experiments with Varying Sizes of HF Points}

In these experiments, we mainly concern with the following issues: 1) the performance of GAN-MDF; 2) the robustness for varying HF sample sizes; 3) the influence on the modeling performance caused by the relatedness between LF and HF responses; and 4) the effectiveness of the supervised-loss trick. In Test-1 to Test-9 ({\it resp.} the practical application problem Test-10), the number of LF points is set to be 100 ({\it resp.} 1000) times the dimension of inputs in each problem, and the number of HF points range from 20 to 2 ({\it resp.} from 30 to 5).

\begin{table}
	\centering				
	\caption{Average NRMSEs for Test Problems with Varying Sizes of HF Samples}
	\label{tab:bi-nrmse}
	\renewcommand\tabcolsep{1pt}
	%\resizebox{\linewidth}{!}{
		\begin{tabular}{cc|cccccc}
			\hline
			Problem  &  $I_L$/$I_H$ & GAN-MDF& HR & H-kriging & LS-MFS & Co-RBF & pGAN\\
			\hline
			Test-1 & 100/5 & {\bf 0.4752} & 1.0000 & 0.9637 & 0.6293 & $0.53\times 10^{2}$ & 1.1123\\
			(1 dim.) & 100/4 & {\bf 0.6361} & 0.6395 & 0.9334 & 0.7470 & $0.72\times 10^{2}$ & 1.4378\\
			& 100/3 & {\bf 0.5308} & 1.0000 & 1.2285 & 0.7412 & 1.1392 & 1.2122\\
			& 100/2 & {\bf 0.6129} & 0.6270 & 1.1461 & 0.9851 & $0.17\times 10^{4}$ & 1.5011\\
			\hline
			Test-2 & 100/5 & {\bf 0.3099} & 0.8233 & 0.3796 & 0.3777 & $0.28\times 10^{2}$ & 2.3677\\
			(1 dim.) & 100/4 & {\bf 0.2542} & 0.8611 & 0.3383 & 0.3603 & $0.16\times 10^{3}$ & 1.7172\\
			& 100/3 & {\bf 0.3151} & 0.8507 & 0.3612 & 0.3156 & 3.4508 & 1.9997\\
			& 100/2 & {\bf 0.2217} & 0.8814 & 0.3449 & 0.3220 & $0.95\times 10^{2}$ & 1.4814\\
			\hline
			{Test-3\tnote{1}} & 100/5 & {\bf 0.9622} & 0.9663 & $1.05\times 10^{2}$ & 1.0426 & $0.79\times 10^{5}$ & 0.9997\\
			(1 dim.) & 100/4 & {\bf 0.9707} & 0.9997 & $0.26\times 10^{2}$ & 1.1405 & $0.20\times 10^{5}$ & 0.9998\\
			& 100/3 & {\bf 0.9429} & 1.0566 & $0.63\times 10^{2}$ & 1.2936 & $0.12 \times 10^{2}$ & 1.0469\\
			& 100/2 & {\bf 0.9872} & 1.0589 & $0.26\times 10^{2}$ & 2.6481 & $0.71\times 10^{8}$ & 2.3358\\
			\hline
			{Test-4} & 100/5 & 0.6705 & 1.0081 & 1.0730 & {\bf 0.6409} & $0.41\times 10^{2}$ & 1.8258\\
			(1 dim.) & 100/4 & {\bf 0.6715} & 1.0167 & 1.2101 & 0.6875 & $0.69\times 10^{1}$ & 1.5488\\
			& 100/3 & {\bf 0.6017} & 1.0485 & 0.9958 & 1.0114 & 0.8029 & 1.5496\\
			& 100/2 & {\bf 0.6492} & 1.0000 & 1.4776 & 0.8254 & 1.7581 & 1.2809\\
			\hline
			Test-5 & 200/20 & {\bf 0.4964} & 0.8172 & 0.8429 & 0.5975 & $0.10\times 10^{2}$ & 3.9884\\
			(2 dim.) & 200/15 & {\bf 0.4607} & 0.8297 & 0.8188 & 0.5983 & 3.4908 & 4.2935\\
			& 200/10 & {\bf 0.5550} & 0.8634 & 0.8436 & 0.8802 & 0.8236 & 3.6395\\
			& 200/5 & {\bf 0.5713} & 0.9432 & 0.8468 & 0.7371 & 1.3031 & 3.2920\\
			\hline
			Test-6 & 600/20 & 0.3562 & 0.4832 & 0.5172 & {\bf 0.3195} & $0.59\times 10^{2}$ & 0.4974\\
			(6 dim.) & 600/15 & 0.4288 & 0.5303 & 0.5262 & {\bf 0.3455} & $0.58\times 10^{2}$ & 0.5366 \\
			& 600/10 & {\bf 0.3407} & 0.4852 & 0.5083 & 0.4057 & $0.63\times 10^{2}$ & 0.4512\\
			& 600/5 & {\bf 0.3260} & 0.4923 & 0.5719 & 0.4840 & $0.34\times 10^{3}$ & 0.4662\\
			\hline
			Test-7 & 800/20 & {\bf 0.4146} & 0.4515 & 1.0306 & 0.7783 & 0.5432 & 0.4881\\
			(8 dim.) & 800/15 & 0.4457 & {\bf 0.4455} & 1.1618 & 0.6485 & 0.5889 & 0.5207\\
			& 800/10 & {\bf 0.4658} & 0.4775 & 1.0104 & 1.7706 & 1.4747 & 0.4922\\
			& 800/5 & {\bf 0.4270} & 0.5331 & 1.0011 & 1.0540 & $0.15\times 10^{2}$ & 0.4936\\
			\hline
			Test-8 & 2000/20 & {\bf 0.3036} & 0.3457 & 0.4515 & 3.1276 & 2.5825 & 1.3327 \\
			(20 dim.) & 2000/15 & {\bf 0.2975} & 0.3602 & 0.4604 & $0.14\times 10^{2}$ & 2.6234 & 3.2627\\
			& 2000/10 & {\bf 0.3140} & 0.4231 & 0.4959 & 4.6967 & $0.86\times 10^1$
			& 1.6758\\
			& 2000/5 & {\bf 0.3002} & 0.3773 & 0.4655 & $0.13\times 10^{2}$ & $0.11\times 10^2$ & 1.7744\\
			\hline
			Test-9 & 3000/20 & {\bf 0.2797} & 0.3971 & 0.4292 & $0.22\times 10^{3}$ & $0.42\times 10^{12}$ & 0.4483\\
			(30 dim.) & 3000/15 & {\bf 0.2683} & 0.4333 & 0.4242 & $0.21\times 10^{3}$ & $0.31\times 10^{12}$  & 0.3484\\
			& 3000/10 & {\bf 0.3037} & 0.4052 & 0.4369 & $0.15\times 10^{3}$ & $0.12\times 10^{9}$ & 0.3644 \\
			& 3000/5 & {\bf 0.2880} & 0.4884 & 0.4386 & $0.65\times 10^{2}$ & $0.29\times 10^{7}$ & 0.3563\\
			\hline
			Test-10 & 2000/30 & {\bf 0.3222} & 0.3571 & 0.9931 & 0.3539 & 0.3539  & 3.4571\\
			(2 dim.) & 2000/25 & {\bf 0.3396} & 0.3600 & 0.9917 & 0.3841 & 0.3539 & 6.0141\\
			& 2000/20 & {\bf 0.3197} & 0.3626 & 0.9923 & 0.3722 & 0.3502 & 5.8077\\
			& 2000/15 & {\bf 0.3318} & 0.4305 & 0.9933 & 0.4606 & 0.3459 & 4.0120\\
			& 2000/10 & {\bf 0.3203} & 0.4744 & $0.54\times 10^{4}$ & 0.4211 & $0.17\times 10^{6}$ & 6.0291\\
			& 2000/5 & {\bf 0.3354} & 0.4263 & $0.48\times 10^{4}$ & 1.1511 & $0.31\times 10^{5}$ & 3.9890\\
			\hline
	\end{tabular}					
\end{table}

\begin{figure*}[htbp]
	\begin{minipage}[t]{0.332\linewidth}
		\centerline{\includegraphics[width=\linewidth]{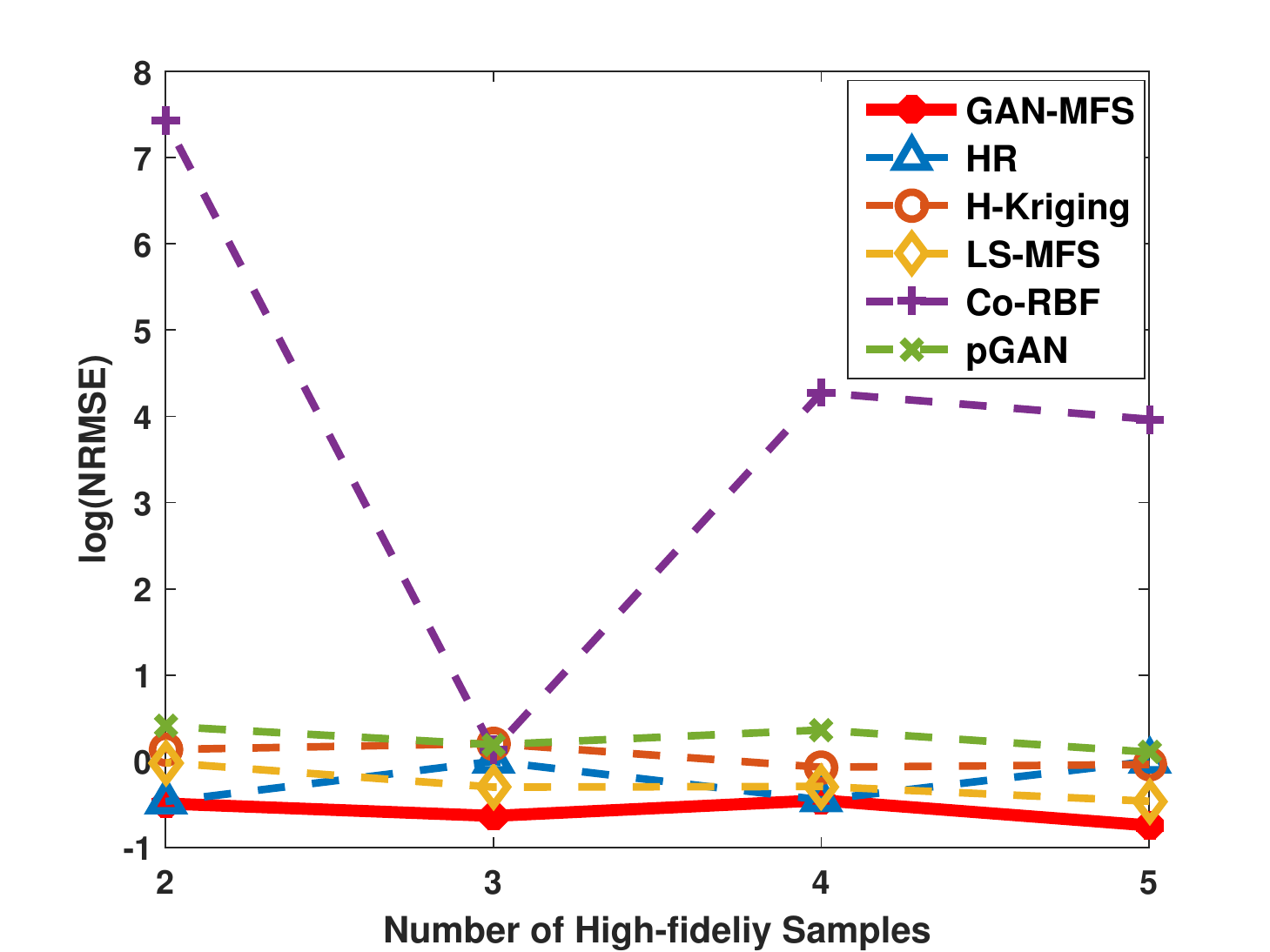}}
		\centerline{(a) Test-1}
	\end{minipage}
	%\hfill	
	\begin{minipage}[t]{0.332\linewidth}
		\centerline{\includegraphics[width=\linewidth]{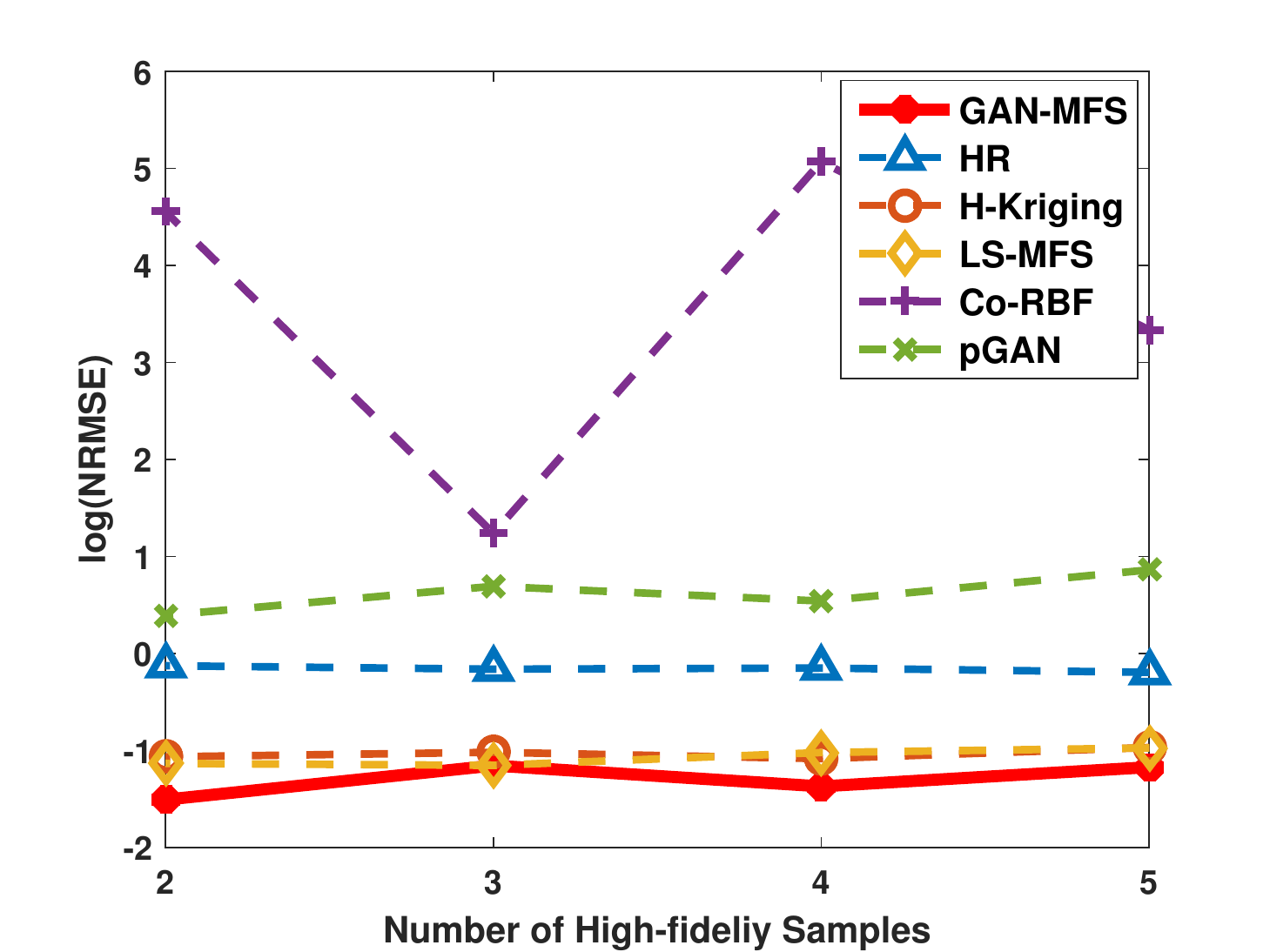}}
		\centerline{(b) Test-2}
	\end{minipage}			
	%\hfill
	\begin{minipage}[t]{0.332\linewidth}
		\centerline{\includegraphics[width=\linewidth]{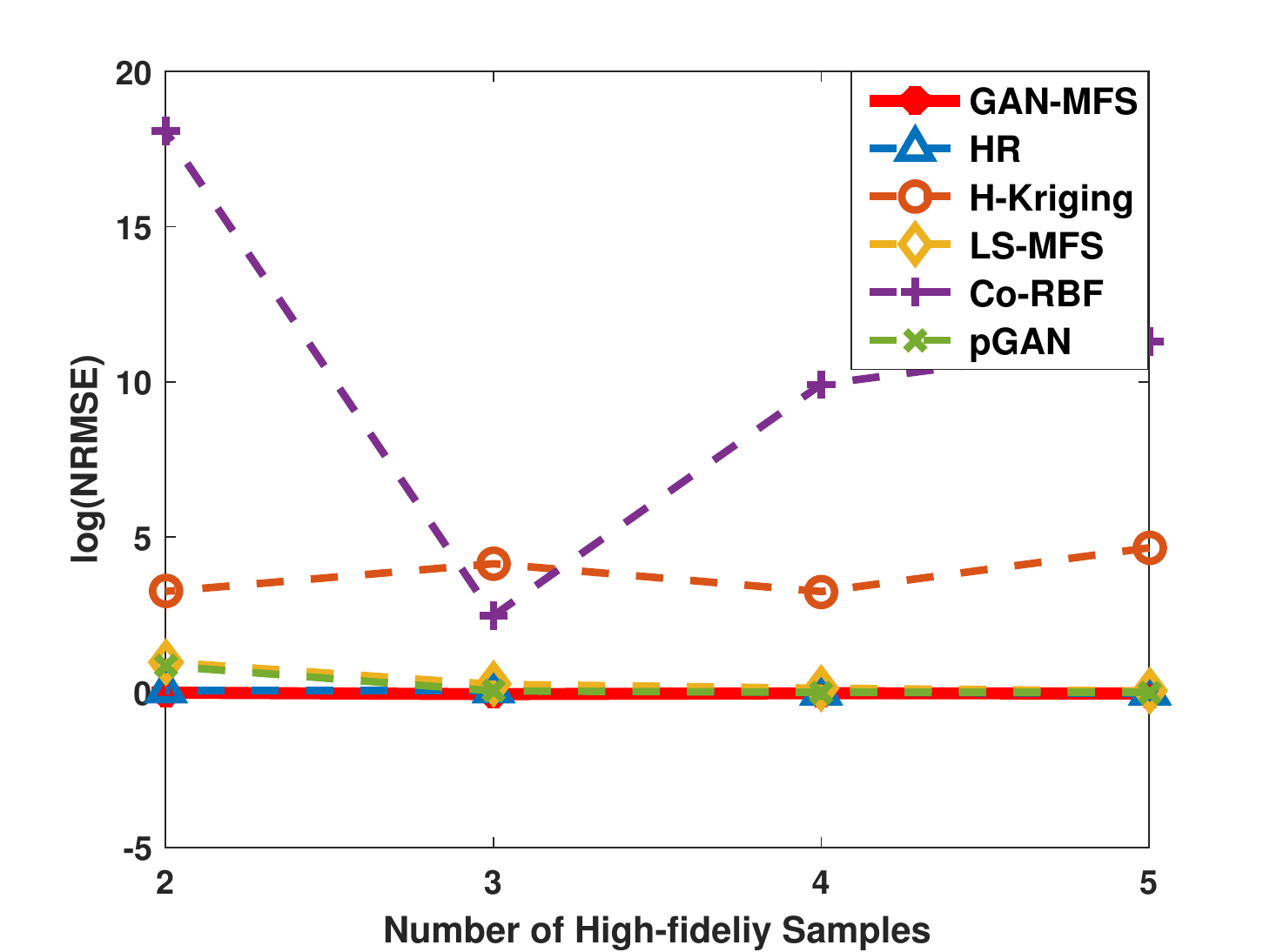}}
		\centerline{(c) Test-3}
	\end{minipage}
	\begin{minipage}[t]{0.332\linewidth}
		\centerline{\includegraphics[width=\linewidth]{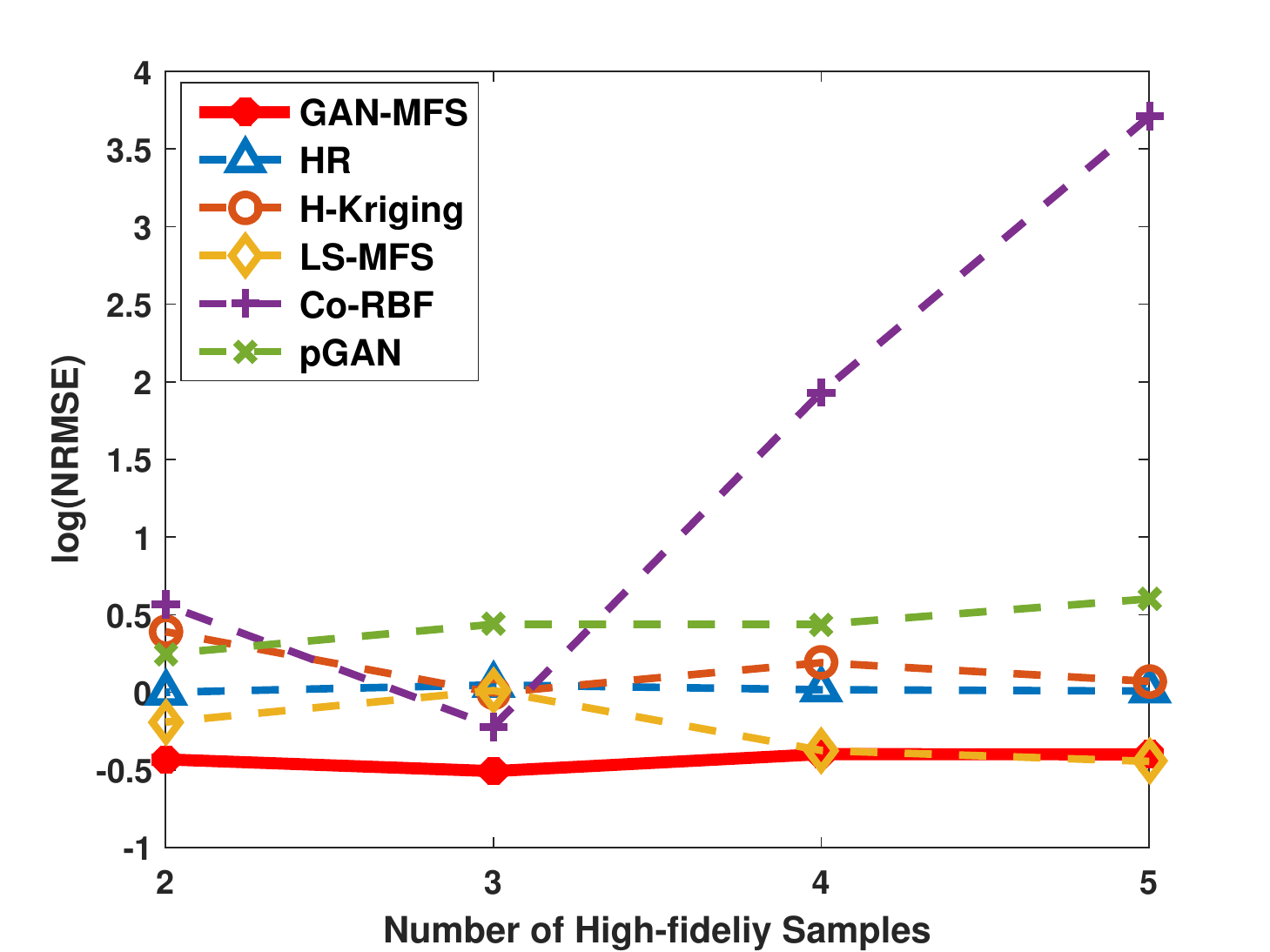}}
		\centerline{(d) Test-4}
	\end{minipage}
	\begin{minipage}[t]{0.332\linewidth}
		\centerline{\includegraphics[width=\linewidth]{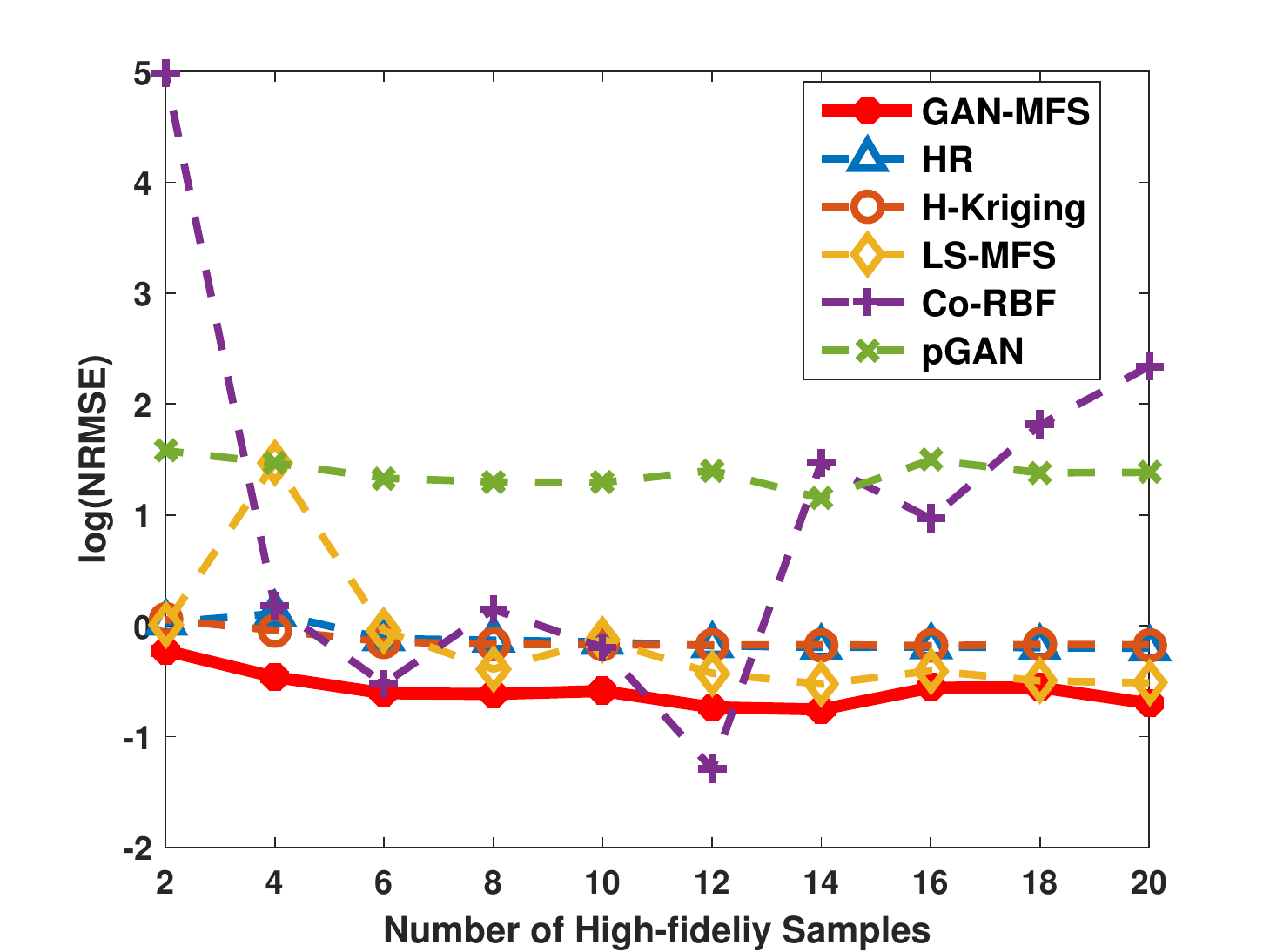}}
		\centerline{(e) Test-5}
	\end{minipage}
	\begin{minipage}[t]{0.332\linewidth}
		\centerline{\includegraphics[width=\linewidth]{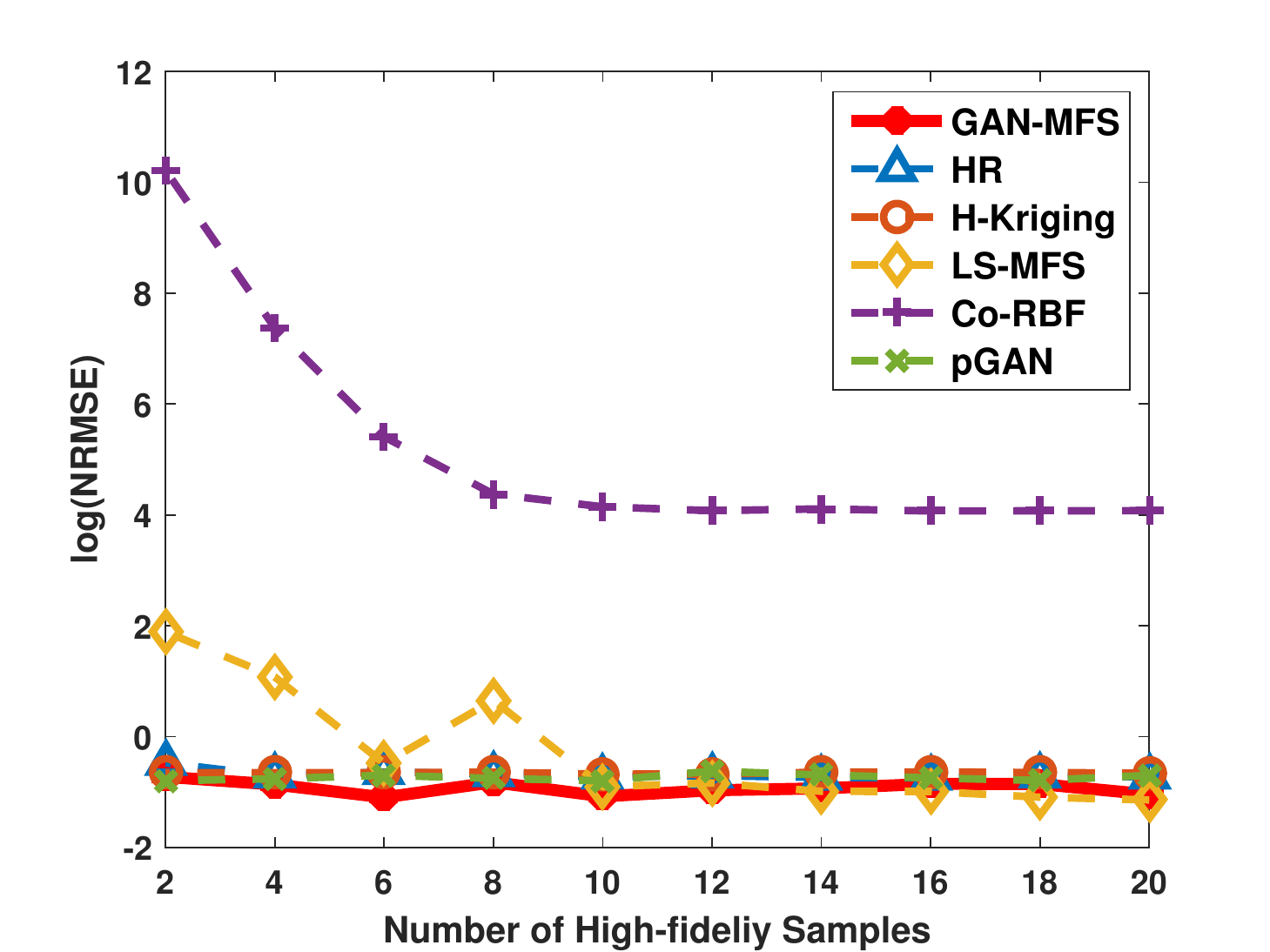}}
		\centerline{(f) Test-6}
	\end{minipage}
	\begin{minipage}[t]{0.332\linewidth}
		\centerline{\includegraphics[width=\linewidth]{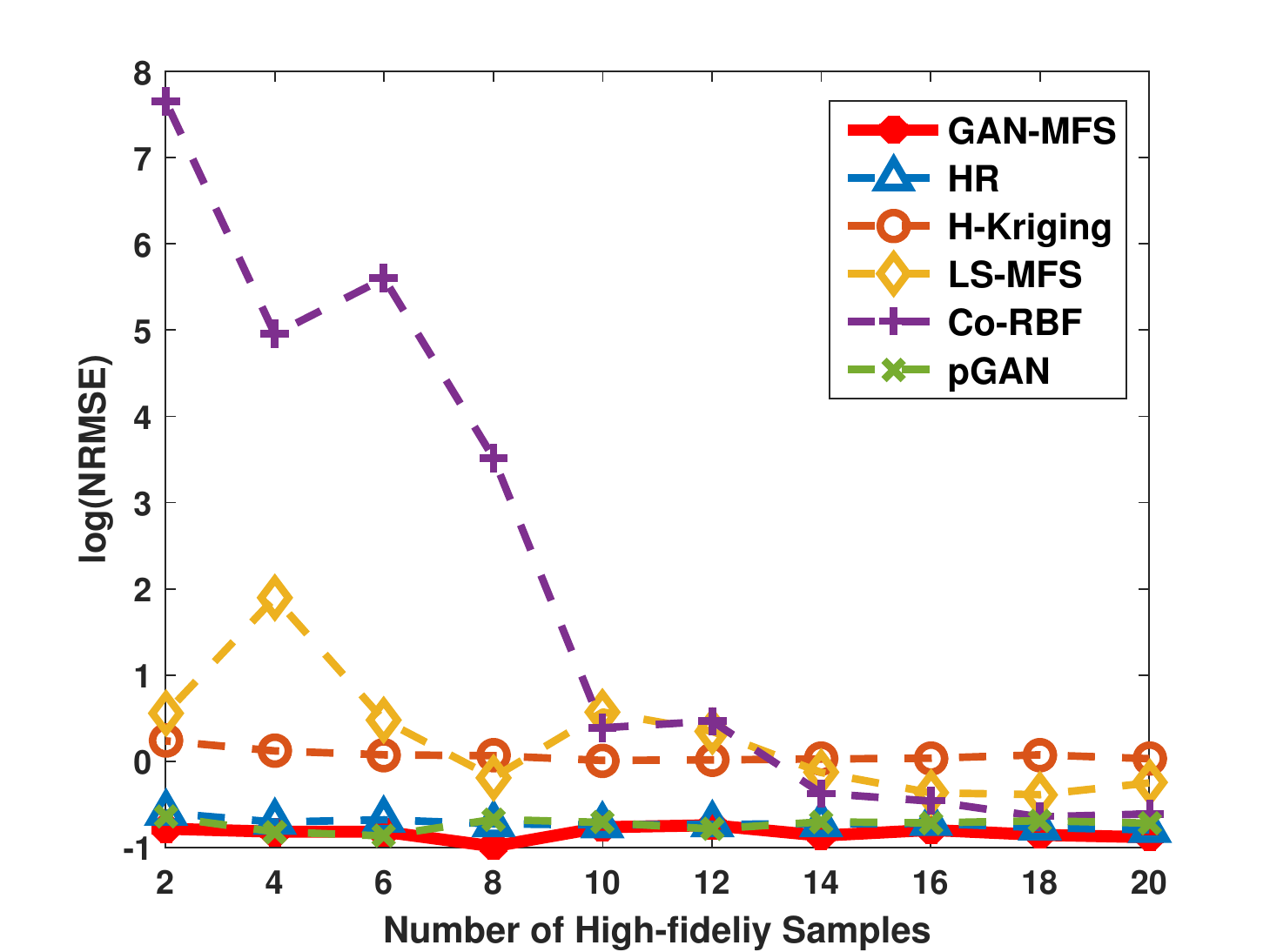}}
		\centerline{(g) Test-7}
	\end{minipage}
	\begin{minipage}[t]{0.332\linewidth}
		\centerline{\includegraphics[width=\linewidth]{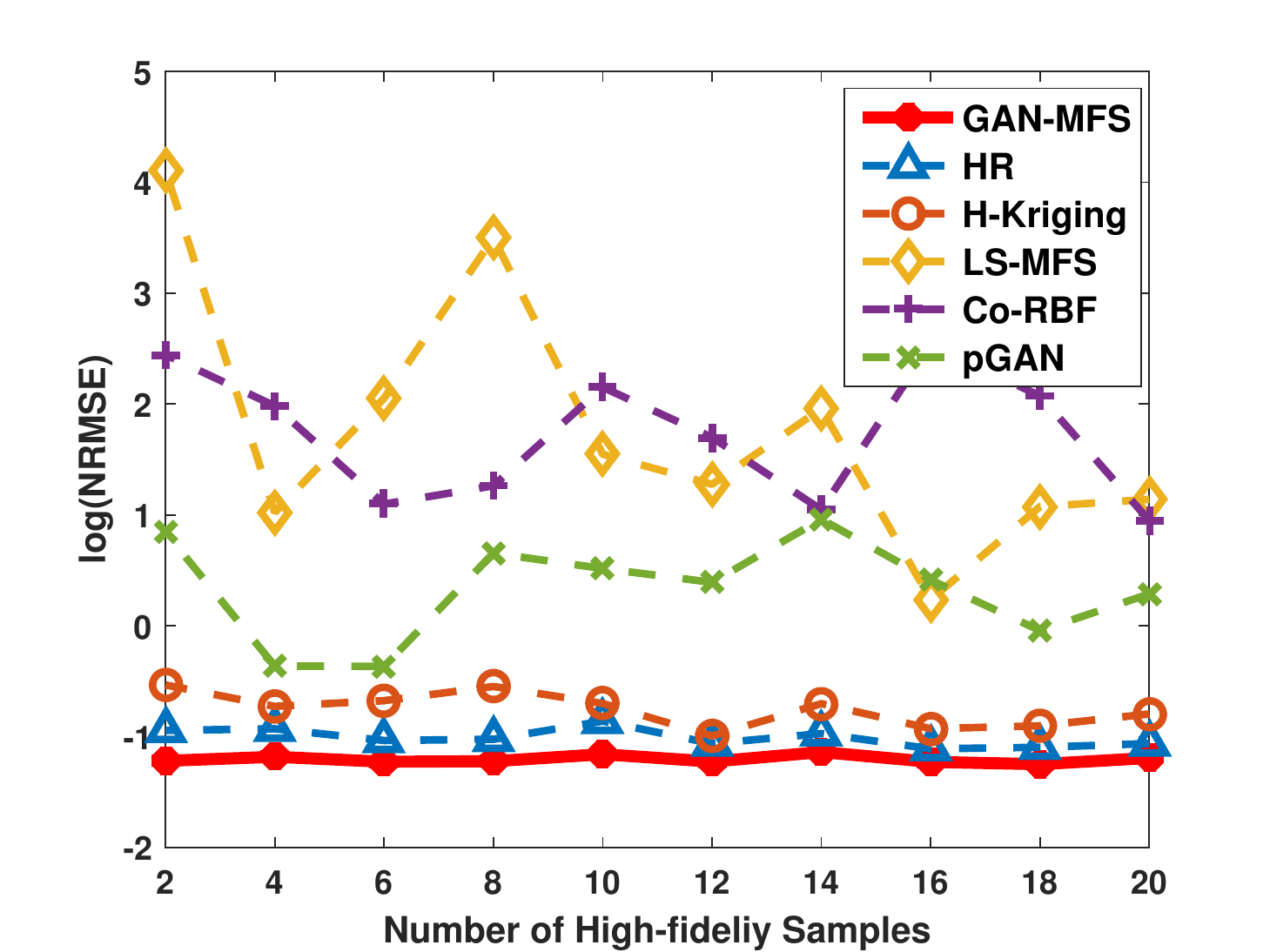}}
		\centerline{(h) Test-8}
	\end{minipage}
	\begin{minipage}[t]{0.332\linewidth}
		\centerline{\includegraphics[width=\linewidth]{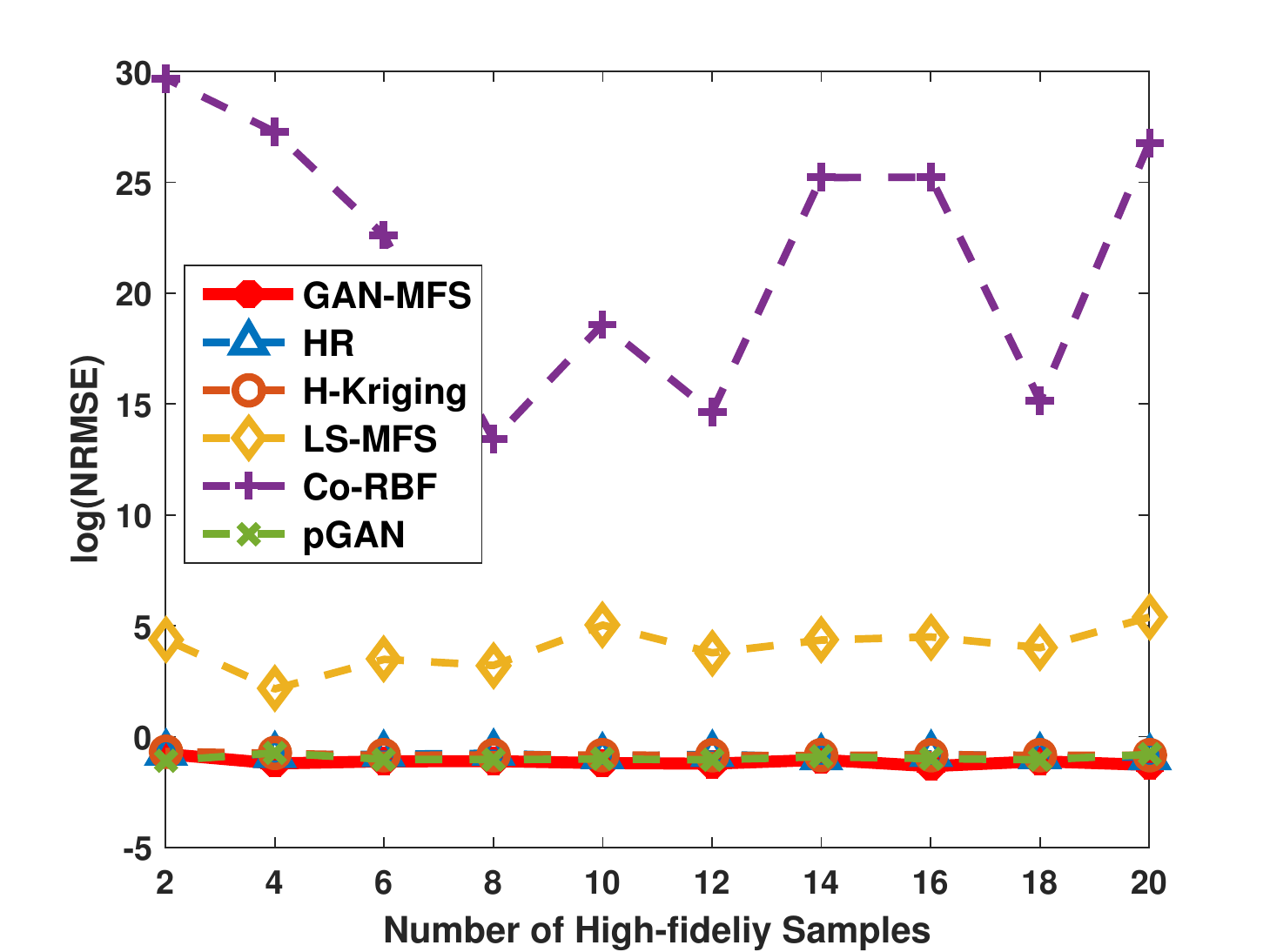}}
		\centerline{(i) Test-9}
	\end{minipage}

	\begin{center}
	\begin{minipage}[t]{0.332\linewidth}
		\centerline{\includegraphics[width=\linewidth]{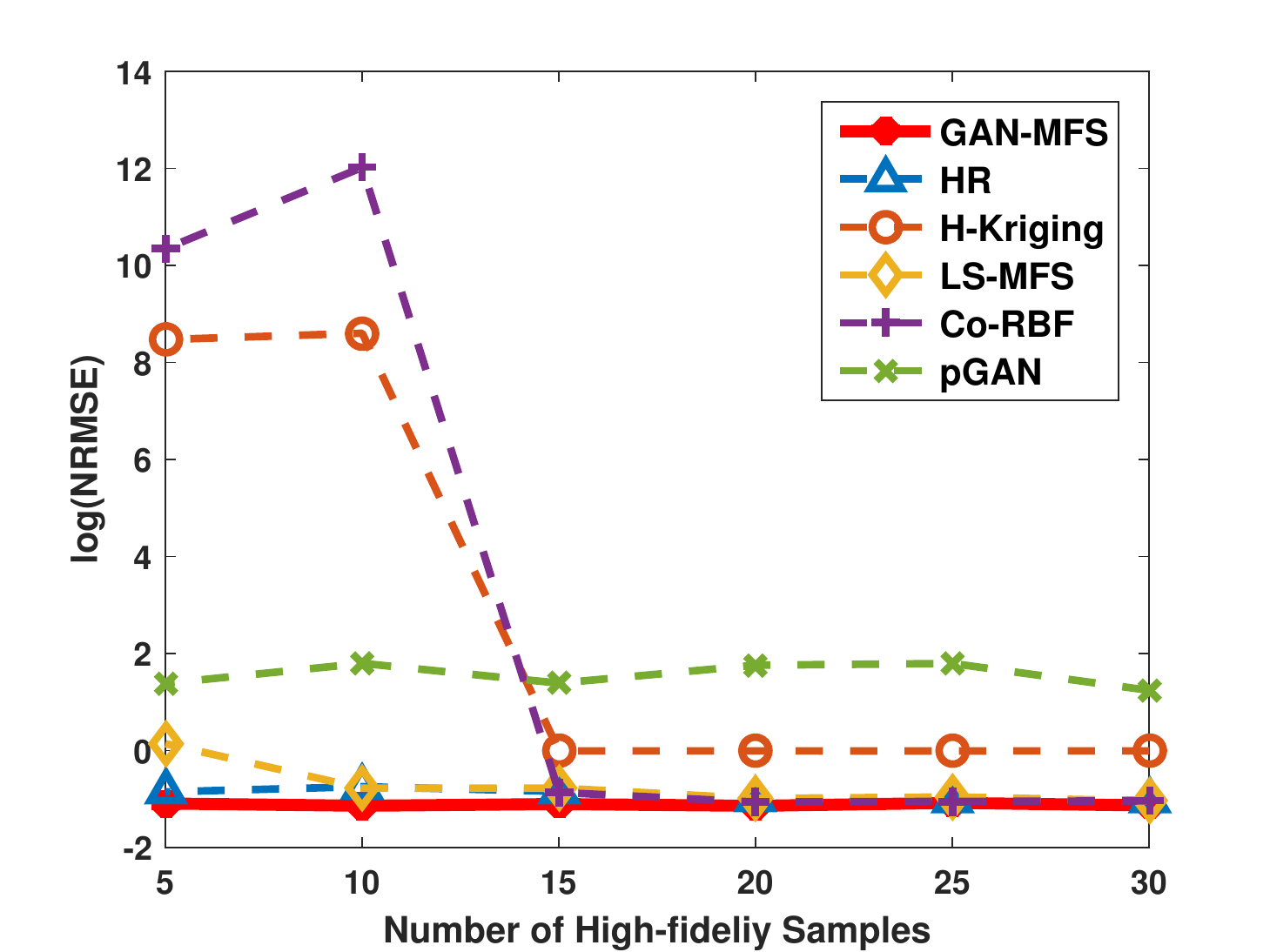}}
		\centerline{(j) Test-10}
	\end{minipage}
	\end{center}

	\caption{The curves of $\log({\rm NRMSE})$ for varying HF sample sizes in benchmark problems}
	\label{fig:bi-nrmse}
\end{figure*}

\begin{figure*}[t]
	\begin{minipage}[t]{0.332\linewidth}
		\centerline{\includegraphics[width=\linewidth]{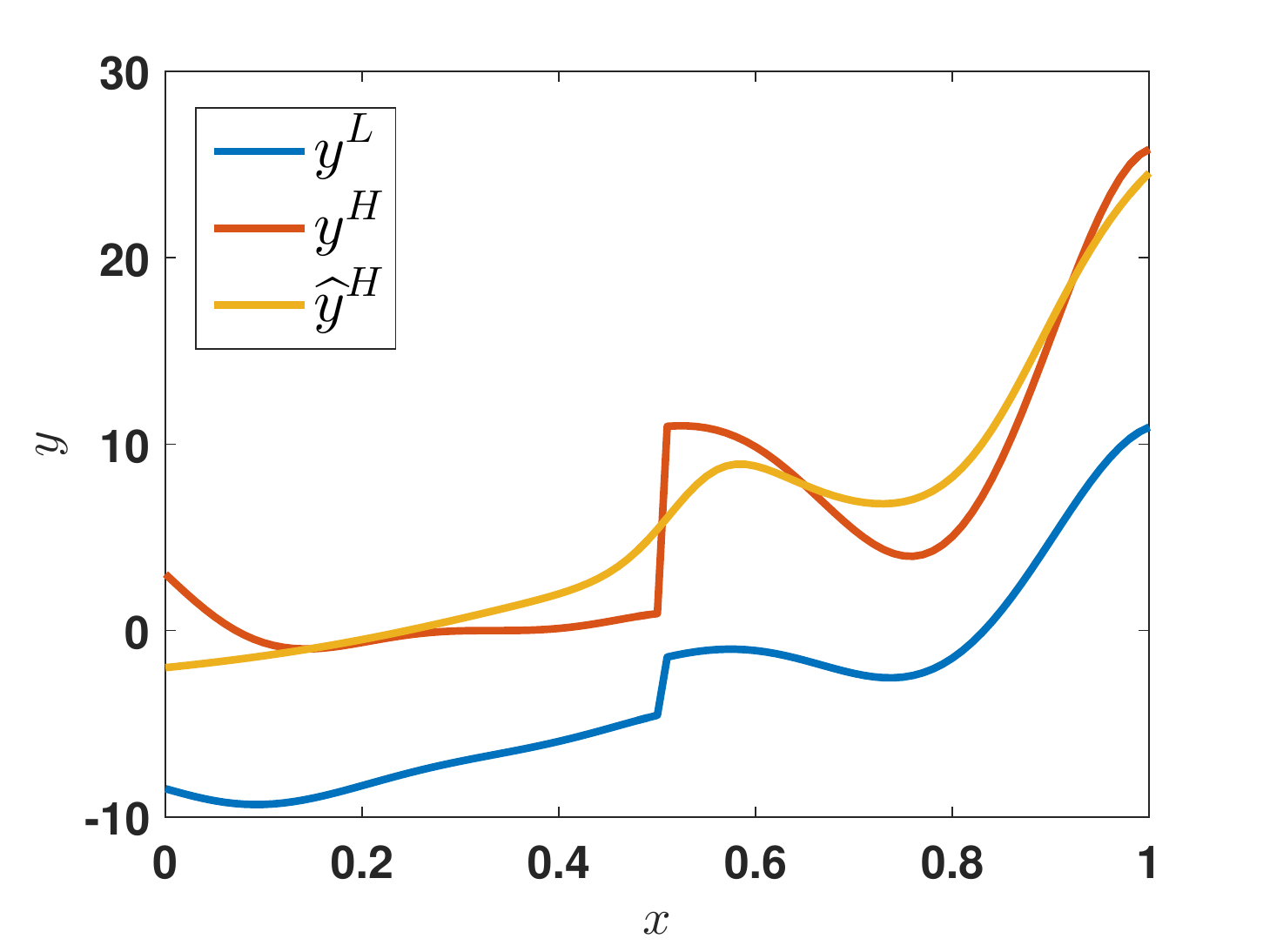}}
		\centerline{(a) Test-1}
	\end{minipage}
	%\hfill	
	\begin{minipage}[t]{0.332\linewidth}
		\centerline{\includegraphics[width=\linewidth]{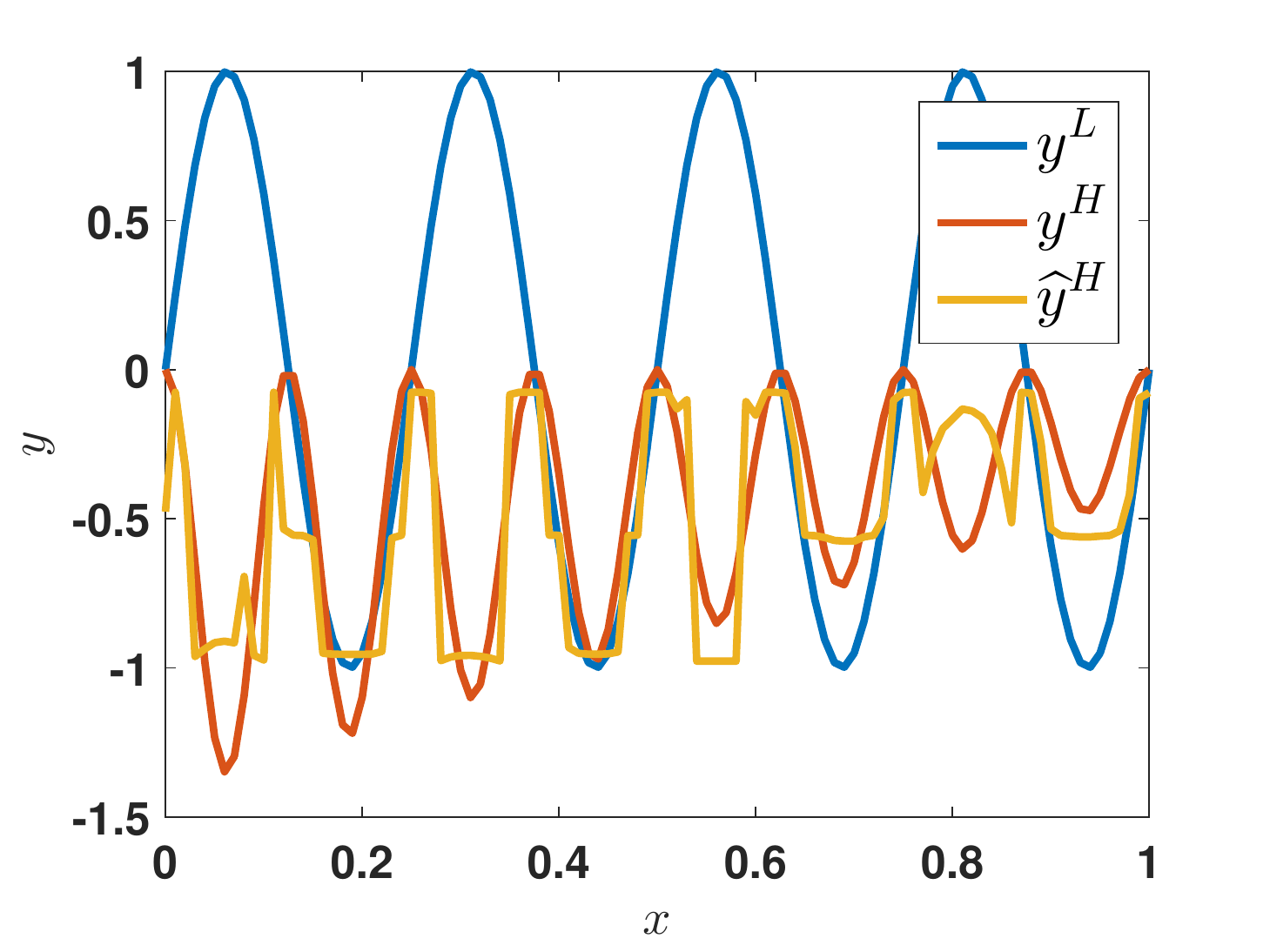}}
		\centerline{(b) Test-2}
	\end{minipage}			
	%\hfill
	\begin{minipage}[t]{0.332\linewidth}
		\centerline{\includegraphics[width=\linewidth]{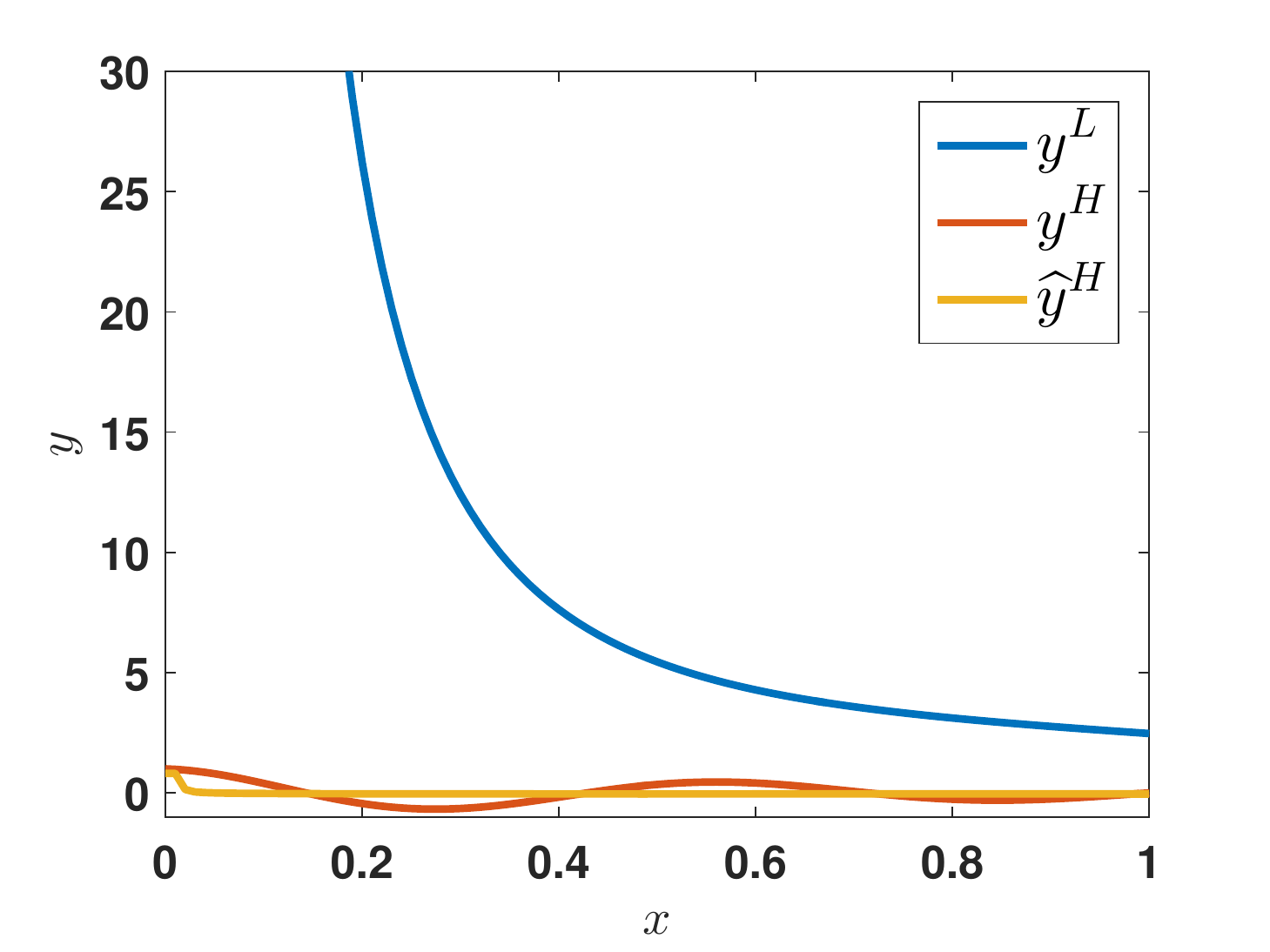}}
		\centerline{(c) Test-3}
	\end{minipage}
	\begin{center}
	\begin{minipage}[t]{0.332\linewidth}
		\centerline{\includegraphics[width=\linewidth]{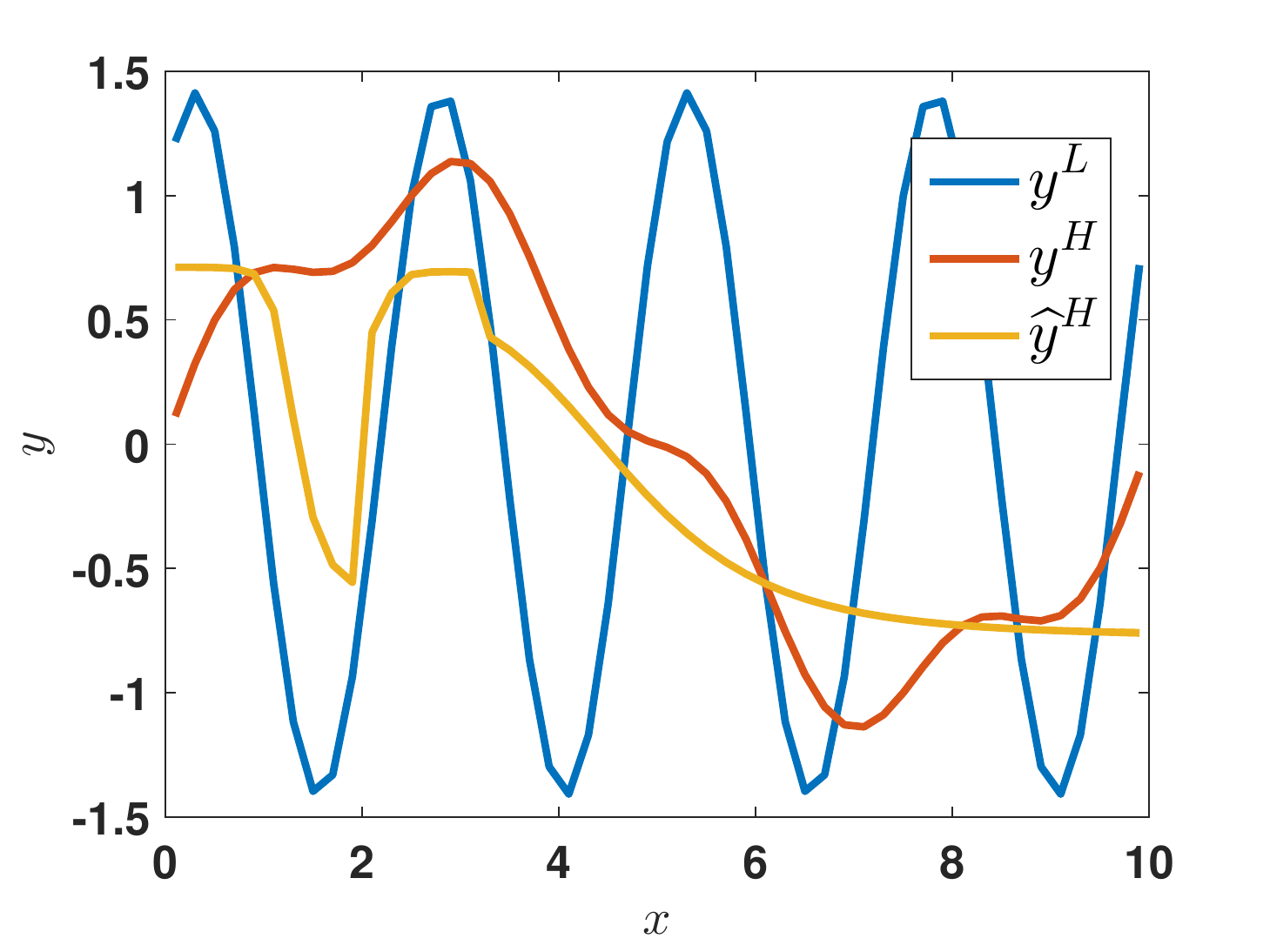}}
		\centerline{(d) Test-4}
	\end{minipage}
	\begin{minipage}[t]{0.332\linewidth}
		\centerline{\includegraphics[width=\linewidth]{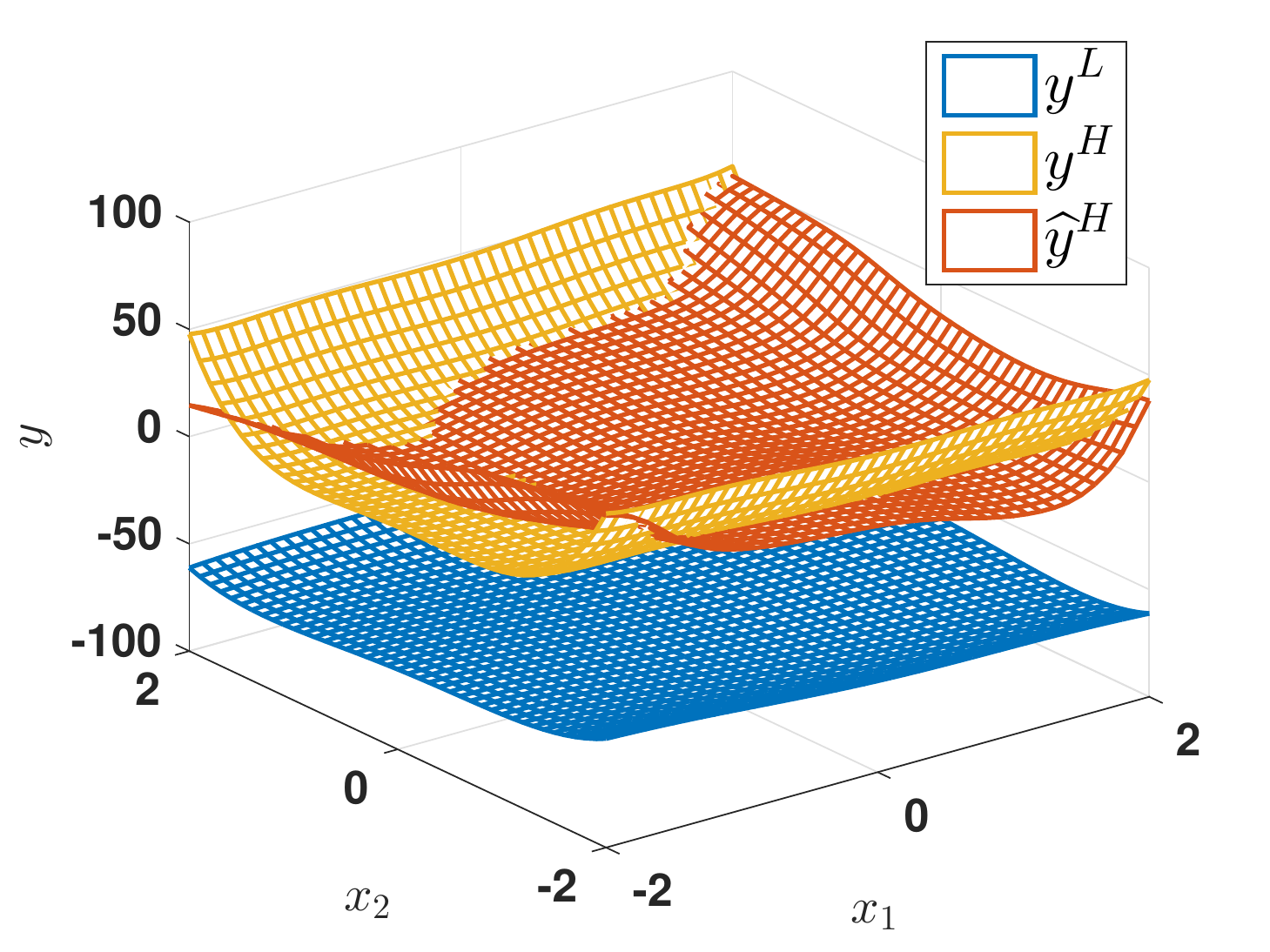}}
		\centerline{(e) Test-5}
	\end{minipage}
    \end{center}

	\caption{The illustration of GAN-MFS modeling results based on {\it five} HF samples in Test-1, Test-2, Test-3 and Test-4 and {\it twenty} HF samples in Test-5}
	\label{fig:mr}
\end{figure*}

As shown in Tab. \ref{tab:bi-nrmse}and Fig. \ref{fig:bi-nrmse}, the proposed GAN-MDF outperforms the state-of-the-art methods with varying sizes of HF points in most cases and especially in the case that there are very few HF samples. Taking advantage of adversarial training, GAN-MDF is less sensitive to the HF sample sizes than the other methods. Namely, GAN-MDF has a higher robustness  when there are very few HF samples in practice. Moreover, GAN-MDF provides a more stable performance than the other models for the testing problems with different functional relations between LF and HF responses. Although pGAN has the same structure as GAN-MDF, the absence of supervised-loss trick causes the former to perform worse than the latter, which implies that the supervised-loss trick plays a key role in the adversarial training of GAN-MDF. In Fig. \ref{fig:mr}, we illustrate the modeling results of the GAN-MDF trained by using {\it five} HF samples in benchmark problems including Test-1, Test-2, Test-3 and Test-4; and {\it twenty} HF samples in Test-5, which show that GAN-MDF approximate the HF responses with high robustness and a satisfactory accuracy.

\subsection{Experiments with Varying Sizes of LF Points}

Here we conduct several experiments to explore how the number of LF samples affects the performances of GAN-MDF. The number of LF samples ranges from 100 to 10 ({\it resp.} from 1000 to 100) times of the dimention of inputs in problem Test-1 to Test-9 ({\it resp.} Test-10), while the number of HF points is set to be 5 for Test-1 to Test-4, 30 for Test-10 and 20 for the rest of benchmark problems. Each experiment is repeated for {\it ten} times and the average NRMSEs are shown in Tab. \ref{tab:nrmse-lf}:

\begin{table}[htbp]
	\renewcommand\tabcolsep{8pt}
	\centering
	\caption{Average NRMSEs of GAN-MFS as the number of LF samples varies}\label{tab:nrmse-lf}
	
	\begin{threeparttable}
		\begin{tabular}{c|ccccc}
			\hline
			\diagbox[width=7em,trim=l]{Problem}{$I_L$} & {$100d$\tnote{1}} & $80d$ & $60d$ & $40d$ & $20d$ \\
			\hline
			Test-1 & 0.4752 & 0.4871 & 0.5275 & 0.4488 & 0.5768\\
			Test-2 & 0.3099 & 0.3044 & 0.3985 & 0.3160 & 0.3058\\
			Test-3 & 0.9622 & 0.9170 & 0.8861 & 0.8684 & 0.8261\\
			Test-4 & 0.6705 & 0.6892 & 0.6574 & 0.6129 & 0.6431\\
			Test-5 & 0.4964 & 0.6176 & 0.4439 & 0.5655 & 0.7156\\
			Test-6 & 0.3562 & 0.3977 & 0.4375 & 0.4126 & 0.4660\\
			Test-7 & 0.4146 & 0.4482 & 0.4426 & 0.4286 & 0.4609\\
			Test-8 & 0.3036 & 0.3004 & 0.2975 & 0.3010 & 0.3066\\
			Test-9 & 0.2797 & 0.3279 & 0.2993 & 0.2839 & 0.2995\\
			\hline
			\hline
			\diagbox[width=7em,trim=l]{Problem}{$I_L$} & $1000d$ & $800d$ & $600d$ & $400d$ & $200d$ \\
			\hline
			Test-10 & 0.3222 & 0.3222 & 0.3221 & 0.3223 & 0.3223\\
			\hline
			
		\end{tabular}
		
		\begin{tablenotes}
			\footnotesize
			\item[1] {We denote "100d" as 100 times of the dimension of inputs.}
		\end{tablenotes}
	\end{threeparttable}
\end{table}			

As the number of LF samples reduces gradually, the benchmark problems can be roughly divided into two classes. In Test-2, Test-4, Test-8, Test-9 and Test-10, the average NRMSEs and SMAPEs hold steady while $I_L$ decreases. Meanwhile, in Test-1, Test-5, Test6, and Test-7, the experimental performances go slightly worse as $I_L$ changes, which means in these benchmark problems the GAN-MDF model is more sensitive to the number of LF points. It is likely that in these problems the functions are complex and the HF points are in such shortage that the model rely on the LF points more. In Test-3, the model performance improves as $I_L$ varies, which means the LF samples cannot provide enough information about HF functions and sometimes too much LF samples have a reverse effect. Since Test-3 is an isolated case, it cannot be representative of other problems.

\section{Conclusion}

In this paper, we propose a generative adversarial network for multi-fidelity data fusion (GAN-MDF) problems in digital twins. The generator of GAN-MDF consists of the LF and the HF blocks. The former captures the LF features from inputs and then encodes them into the subsequent network structure; and the latter produces the accurate approximation of HF responses. The discriminator identifies whether the HF-block output is an observation of the sample distribution. When the discriminator is trained to be incapable of identifying the output of HF block or a real HF response, the generator can produce an accurate approximation of the HF response. We also introduce the supervised-loss tricks to enhance the stability of GAN-MDF's training process. The experimental results validate the proposed GAN-MDF, and show that 1) GAN-MDF outperforms the state-of-the-art methods including the co-RBF method, the H-kriging method, the LS-MFS method and the HR method in most cases with different functional relations between LF and HF responses; 2) it has high robustness for varying HF training sample sizes even when there are very few HF samples;  and 3) The supervised-loss trick plays an essential role in the training of GAN-MDF.

Moreover, we conduct more experiments to study the impacts of varying sizes of LF samples on the performances. In this setting the ten benchmark problems can be roughly divided into two classes, one is not very sensitive to the number of LF samples and the other is on the contrary. It appears that the latter class of problems are rather complex and the models are forced to count on more on these LF points with insufficient HF points. This suggests that when testing problem is simple and clear, $I_L$ does not have to be of a very high number, and vice versa.
In our future works, we will study the theoretical properties of MDF methods, and present some theoretical explanations to the experimental findings.

\end{spacing}

\bibliographystyle{unsrt}
%\setcitestyle{numbers}

\bibliography{GAN-MDF}

\end{document}